\documentclass[letterpaper, 10 pt, conference]{ieeeconf}  
\usepackage{threeparttable}
\usepackage{graphicx}
\usepackage{subfigure}
\usepackage{xcolor}
\usepackage{amssymb}

\IEEEoverridecommandlockouts                              

\title{\LARGE \bf
Exploiting Point-Wise Attention in 6D Object Pose Estimation Based on Bidirectional Prediction
}

\author{Yuhao Yang$^{\ast}$, Jun Wu$^{\ast}$, Yue Wang, Guangjian Zhang and Rong Xiong$^{\dagger}$
\thanks{$^{\ast}$These authors contributed equally to this work.}%
\thanks{$^{\dagger}$Corresponding author
        {\tt\small rxiong@zju.edu.cn}}%
\thanks{Yuhao Yang and Guangjian Zhang are with the School of Artificial Intelligence, Chongqing University of
Technology, Chongqing, 100190, China.}%
\thanks{Jun Wu, Yue Wang, and Rong Xiong are with the College of Control science and Engineering, Zhejiang University,
        Hangzhou, 310027, China.}
\thanks{This work was supported in part by the National Nature Science Foundation of China under Grant 62173293 and in part by the Zhejiang Provincial Natural Science Foundation of China (LD22E050007).}
}

\begin{document}

\maketitle
\pagestyle{empty}  
\thispagestyle{empty} 

\thispagestyle{empty}
\pagestyle{empty}

\begin{abstract}

Traditional geometric registration based estimation methods only exploit the CAD model implicitly, which leads to their dependence on observation quality and deficiency to occlusion.To address the problem,the paper proposes a bidirectional correspondence prediction network with a point-wise attention-aware mechanism. This network not only requires the model points to predict the correspondence but also explicitly models the geometric similarities between observations and the model prior.
Our key insight is that the correlations between each model point and scene point provide essential information for learning point-pair matches. To further tackle the correlation noises brought by feature distribution divergence, we design a simple but effective pseudo-siamese network to improve feature homogeneity. 
Experimental results on the public datasets of LineMOD, YCB-Video, and Occ-LineMOD show that the proposed method achieves better performance than other state-of-the-art methods under the same evaluation criteria. Its robustness in estimating poses is greatly improved, especially in an environment with severe occlusions.

\end{abstract}

\section{INTRODUCTION}

The object 6D pose estimation task is to compute the object’s 3D rotation and 3D translation in the current scene with respect to the canonical coordinates. It is an essential problem in human-robot interaction applications such as augmented reality~\cite{1}, autonomous driving~\cite{2}, and robot manipulation~\cite{4}. 
Unlike category-level or unseen object pose estimation tasks \cite{wang2019normalized,chen2020category},
When tackling the instance-level object pose estimation problem, a CAD model of the target object is generally specified. The model establishes the canonical coordinates, and contains the distinctive features of the target, providing vital prior for estimation. Herein lies one of the key research issues - how to utilize the CAD model for object pose estimation. 

\begin{figure}
	\centering
		\includegraphics[scale=0.85]{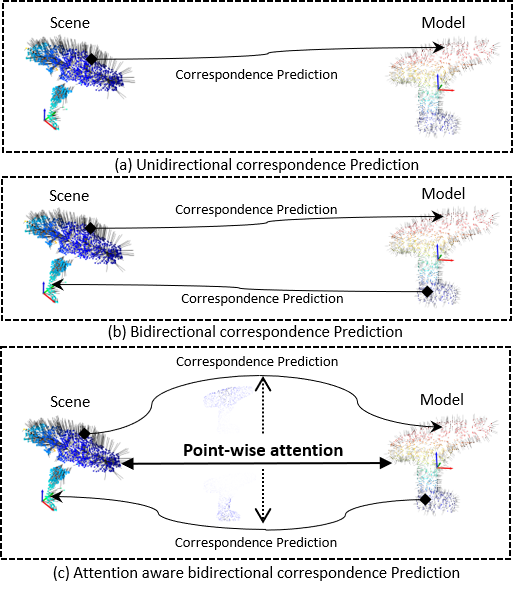}
  \vspace{-4mm}
	\caption{\textbf{Illustration of our idea.} We show the difference between unidirectional match prediction methods (a), bidirectional match prediction methods (b), and our point-wise attention bidirectional match prediction method (c).}
	\label{FIG:IDEA}
 \vspace{-0.65cm}
\end{figure}

Some approaches~\cite{7,li2018deepim} intuitively exploit the CAD model by generating observations of the model under different poses with perspective projection, then comparing the query scene with the generated observations, and optimizing the pose to descend their differences. They directly harness the rendered features from the CAD model, but are rather sensitive to the initial guess and prone to a local minimum because the mapping between the pose and observation is not unimodal. 
On the other hand, some methods leverage the powerful fitting capability of the neural network to extract features for predicting the correspondence between the scene and the model~\cite{34,17}. 
This pipeline adopts ground truth correspondences as supervision, obtaining global solutions to achieve higher accuracy. 
Though the CAD model is explicitly utilized in the registration stage, its point features are not fully exploited in the prediction process, causing dependence on the quality of observation.
When the observations are incomplete or noisy, the global solution could also be affected.
Recently, \cite{43} attempts to further exploit the CAD model by introducing an inverse prediction process that predicts the corresponding scene points for each model point. But they only employ the averaged global features with point-wise features from each point set, disregarding their mutual attention that is obliging for correspondence prediction. 

In this paper, we propose a bidirectional point-wise attention aware network for stable 6D object pose estimation. We adopt two branches to predict the correspondence from scene points to model coordinates and vice versa, and design a geometric attention mechanism to assist the prediction. Our key insight is that the correlations between each model point and scene point provide essential information for learning point-pair matches. The scene points are essentially model points observed in a specified view with noises and occlusions. Since the geometric properties of pointclouds do not vary with changes in viewing perspective, corresponding points in the scene and the model should still have the highest attention response to each other despite the transformation. During training, the known transformation is applied to supervise the learning of attention mechanism, enabling the network to model the point-wise attention during inference. This attention module is then concatenated with the feature vector to predict the correspondence. Experimental results demonstrate that by using the attention mechanism module, the correspondence prediction and pose estimation performance is improved.

We further leverage a simple yet effective pseudo-siamese network (PSN) to obtain point-wise attention. 
Intuitively, the coveted attention could be calculated directly from the features extracted for correspondence prediction~\cite{8}. However, we argue that these features are insufficient for attention awareness. The scene features often include color properties learned from the input RGB image, while the model features do not, which disturbs the mutual correlation computation. Moreover, the scene features and model features generally follow different distributions because they are extracted separately from two branches. We hope that the attention only reflects the geometric similarities between point pairs, rather than being affected by distribution divergences. Therefore, we design an additional pseudo-siamese neural network, which takes both sets of point clouds as input and extracts their features for attention calculation. 

To summarize, the contributions of this paper are mainly as follows:
\begin{itemize} 
\item To exploit the CAD model for stable 6D object pose estimation, we propose a bidirectional match prediction network with global point-wise attention aware mechanism, and prove its effectiveness in improving point pair match learning. 
\item To obtain robust attention, we introduce a simple but effective pseudo-siamese network to discover the similarities between model points and scene points. 
\item We validate our proposed method on public datasets of LineMOD, YCB-Video, and Occ-LineMOD. The experimental results show that our network outperforms state-of-the-art methods in both accuracy and robustness. 
\end{itemize}

\begin{figure*}
	\centering
		\includegraphics[scale=.4]{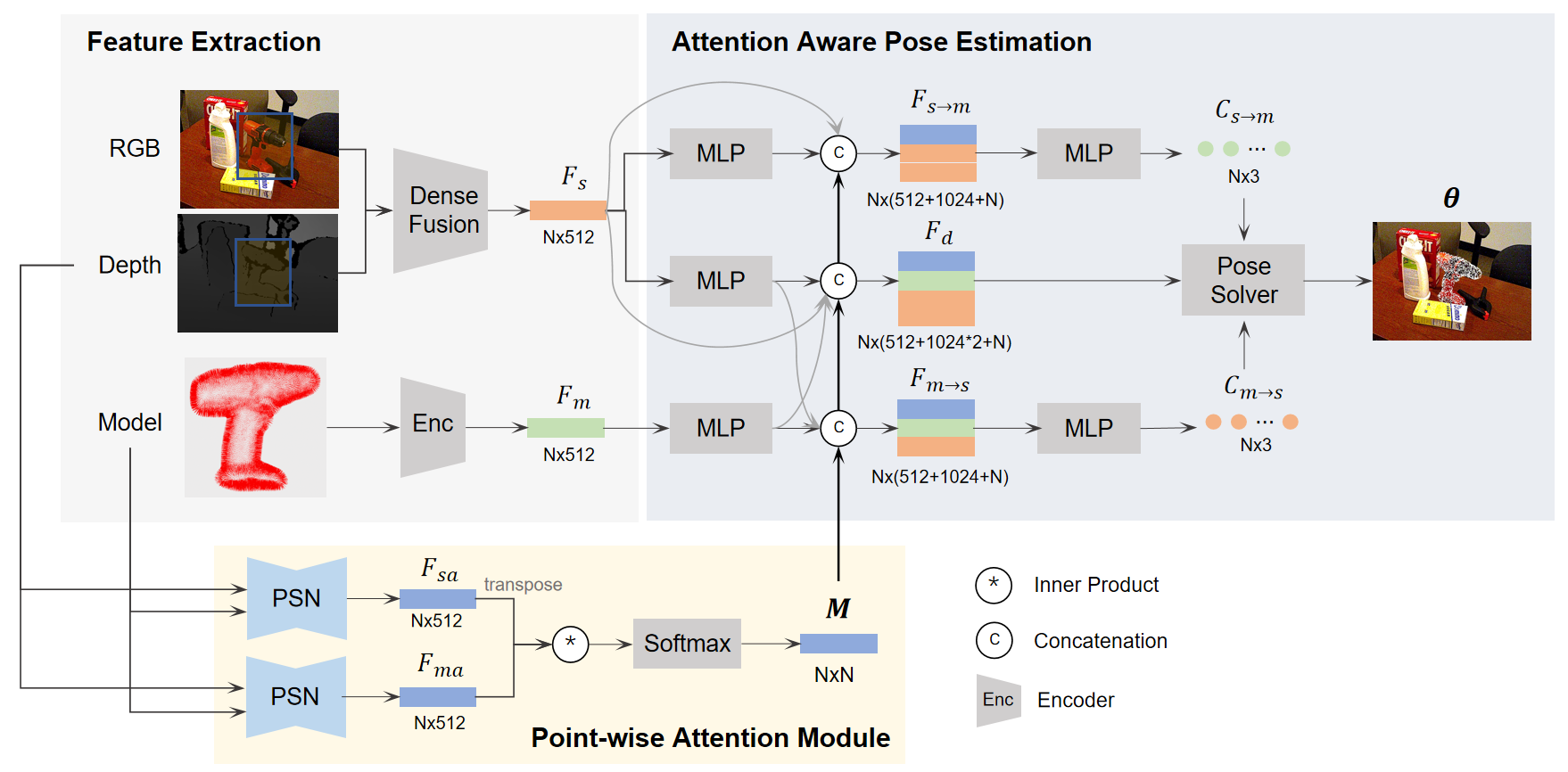}
\vspace{-4mm}  
	\caption{Overview of our proposed method. We propose a point-wise attention module to obtain the correlations between scene points and model points. The attention is concatenated with other learned features to predict bidirectional correspondence and solve poses.}
	\label{fig:network}
\vspace{-0.4cm}
\end{figure*}

\section{Related Works}

\subsection{Utilizing CAD models by comparison methods}
Since the CAD model is available in the instance-level object pose estimation task, early methods directly generate templates by projecting the 3D model with various angles, and estimate the query pose by finding the most similar template image~\cite{24}\cite{gordo2016deep}\cite{gu2010discriminative}.
\cite{7}\cite{hinterstoisser2013model} propose a novel image representation by spreading image gradient orientations and representing the object with a limited set of templates. 
\cite{hodavn2015detection} verify the candidates by matching features in different modalities and associate the approximate poses with each detected template as the initial value for further optimization. 
These methods take advantage of explicitly comparing with the CAD model and are capable of handling textureless objects, but the discretized template generation process leads to less accuracy. 

Recently, some methods adopt graphical rendering techniques to generate pseudo observations in continuous pose space~\cite{manhardt2018deep}\cite{trabelsi2021pose}\cite{yen2021inerf}\cite{10}.
\cite{li2018deepim} exploits a deep learning-based pose refinement network to refine the initial pose iteratively by minimizing the differences between the observed image and the rendered image. 
\cite{wang2021nemo} proposes a pose refinement method using the standard differentiable rendering and learning the texture of a 3D model via contrastive loss.
\cite{iwase2021repose} utilizes differentiable Levenberg-Marquardt optimization to refine the pose by minimizing the distance between the input and rendered image representations.
\cite{tracknet} identify the relative pose given the current observation and a synthetic image rendered from the previous estimates. 

These methods have shown great performance by adopting render and compare refinement as a post-processing step, but they are time-consuming and sensitive to initial guess.

\subsection{Utilizing CAD models by prediction methods}
Instead of explicitly utilizing the generated images from the CAD model, another pipeline requires the network to implicitly learn the correlation between the observation and the model.
With the recent advancements in deep learning, several methods \cite{rad2017bb8}\cite{tekin2018real}\cite{luo20203d}\cite{mitash2018robust} have attempted to 
detect box corners in the RGB image for 3D bounding box estimation.
PVNet \cite{17}, employs farthest point sampling to vote for key points on the target object, predicting the direction vector pointing from each pixel to the projection point using a RANSAC voting strategy to locate the projection point. Some subsequent methods attempt to enhance the precision of correspondence prediction or the robustness of geometric solver~\cite{wen2020robust}\cite{brachmann2019dop}\cite{34}\cite{chen2022epro}\cite{hua2021rede}. 
Recently, some approaches have attempted to include model prior information 
to provide geometric constraints.
\cite{moban} proposes a graph neural network to learn implicit neural representations of the 3D model and presents a dense correspondence matching scheme for visible points. 
BiCo-Net\cite{43} adds an extra branch to predict the correspondence from model points to scene points, achieving higher robustness against occlusion. But they rely on the network to predict the correspondence correctly, ignoring the correlations between each point pair. While we further propose a global attention mechanism to leverage the difference among predicted correspondences.

Besides solving the registration problem from predicted correspondence, many other methods attempt to implicitly utilize the CAD model through direct regression approach~\cite{wang2021gdr}\cite{di2021so}\cite{hu2020single}.
DenseFusion~\cite{11} deploys a dense fusion strategy to fuse color features and geometry features point-wisely, then directly regresses pose from the fused feature.
FFB6D~\cite{12} adds a bidirectional fusion module to fuse the two kinds of features on each encoding layer, bringing more local and global features.
GCCN\cite{8} 
applies a co-attention module to compute the correlations between scene points and model points. But they 
apply the attention map with other features to directly regress the pose, reducing the influence of geometric properties. We show in the experiment section that our design of the pseudo-siamese network and geometric solver enhance the capability of the attention mechanism.

\section{Methods}
\subsection{Overall Network Structure}

The main structure of the network is illustrated in Fig.~\ref{fig:network}. Firstly, 
the interested region of the target object is cropped from the RGB and depth images as the input to the network. 
Then, we follow \cite{43} to extract color and geometric features from the scene observations and fuse them point-wisely, as well as extract the model features.
After feature extraction, we propose a point-wise attention module to model the correlations between the scene points and the model points.
Last, we explain how we exploit the learned attention map for pose estimation.

\subsection{Feature Extraction}
Firstly, we crop the region of interest of the target object from the RGB and depth images as the input. Since segmentation is not the focus of our work, we use ground truth masks as in previous works \cite{11}. 

To extract color features, we follow ConvNext \cite{36} to extract surface texture features from the input RGB image.
Then we randomly sample $N$ points from the scene point cloud obtained from the depth image, and follow PointNet~\cite{39} to extract geometric features, which is further concatenated with their corresponding color embeddings to get the dense point-wise feature vector $F_{s}$.
Besides, for bidirectional prediction, we also sample $N$ points from the model points and extract features $F_m$ from the model following PointNet~\cite{39}.




\subsection{Point-wise Attention Module}
In this module, we consider the effect of global point pair geometric attention on the robustness of the final predicted poses and design the global point-wise attention module as shown in Fig.~\ref{fig:network}. 

\textbf{Pseudo-siamese Network.}
In order to obtain features for building the attention between scene points and model points, existing method~\cite{8} proposes to deploy two PointNets for each point set to extract their features, which are then compared to get their correlation. But these features follow different distributions.
Therefore, we design a pseudo-siamese neural network, which takes both sets of point clouds as input. By doing so, the attention map only reflects the geometric similarities between point pairs, rather than being affected by distribution divergences.

\begin{figure}
	\centering
		\includegraphics[scale=.3]{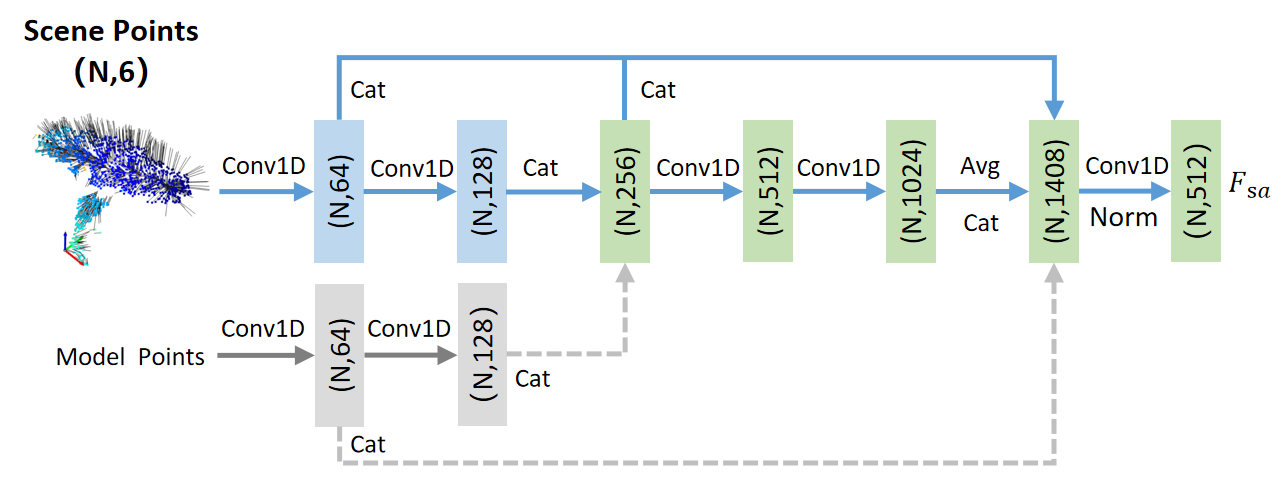}
  \vspace{-4mm}
	\caption{Detailed structure of the proposed pseudo-siamese network. }
	\label{FIG:PSN}
 \vspace{-0.6cm}
\end{figure}

As shown in Fig.~\ref{FIG:PSN}, we input the sampled scene points and their normal vectors $(x,n) \in R^{N*6}$ into the network. 
Then 6 Conv1D layers are adopted to extract their geometric features.
In order to preserve multi-level features, we perform short-circuits to connect the features from top layers to the last layer.
Also, we observe that further concatenating exterior features from the model points could effectively advance the distribution consistency for afterward similarity computation. 
After concatenating the multi-level and exterior features, we obtain a fused feature of size $F \in R^{N*1048}$.
Last, the fused feature is fed to an output Conv1D layer and then normalized to get the final scene point feature $F_{sa} \in R^{N*512}$.
It is the same for the PSN to process model points to get $F_{ma} \in R^{N*512}$, 
except for that the exterior features are from scene points. 
Given $F_{sa}$ and $F_{ma}$, we then apply inner production to take the two feature matrices as input and use the softmax function to generate the attention map M.




\begin{figure}
	\centering
		\includegraphics[scale=.2]{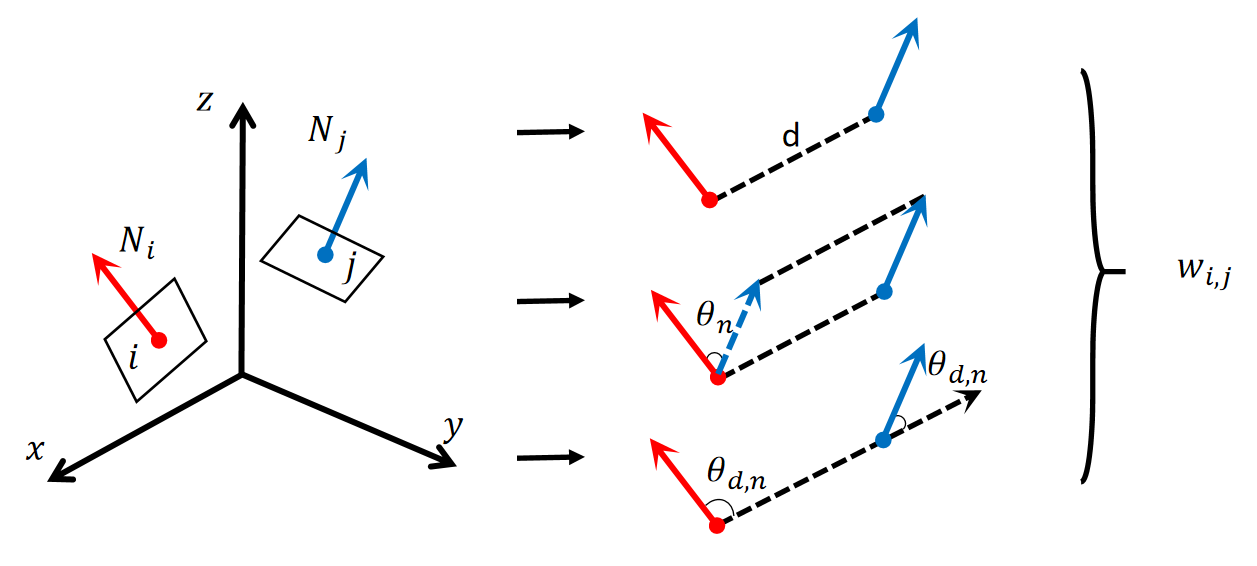}
  \vspace{-4mm}
	\caption{Illustration of point-pair features for attention constraint.}
	\label{FIG:3}
 \vspace{-0.5cm}
\end{figure}

\textbf{PPF constrains.}
To enhance the accuracy and effectiveness of the generated attention maps in focusing on geometric features, a PPF (Point Pair Feature)~\cite{24} constraint term is utilized as supervision to guide the attention maps. 
PPF is an effective way to calculate the relative positions and normal vector directions between point pairs, which enables it to capture the surface invariance and possess tolerance to pose changes. Additionally, the computation of PPF features takes into account the neighborhood information in the point cloud, which is beneficial in handling the case of partial occlusion.
Specifically, all points in the scene point cloud are converted into the canonical coordinate using ground truth poses. Then each point in the transformed point cloud is compared with all points of the model point cloud to calculate the PPF-constrained point-pair features.
As shown in Fig.~\ref{FIG:3}, given the ${i_{th}}$ transformed scene point $(x_i, n_i) \in \mathrm{R}^{N\times 6}$ and the ${j_{th}}$ model point $(x_j, n_j) \in \mathrm{R}^{N\times 6}$
, we calculate the Euclidean distance feature ${d}_{i,j}$, the normal vector angle feature $\theta_{i,j}$, and the distance vector and normal vector angle feature $\theta_{d_{i,j}}$ as follows


\begin{equation}
    {d}_{i,j}=\Vert x_i - x_j \Vert_2
\end{equation}

\begin{equation}
    \theta_{i,j}=\arccos \left(\frac{n_i \cdot n_j}{\left\|n_i\right\|\left\|n_j\right\|}\right)
\end{equation}

\begin{equation}
    \theta_{d_{i,j}}=\arccos \left(\frac{n_i}{\left\|n_i\right\|} \cdot\left(\frac{n_j}{\left\|n_j\right\|}\right)^T \cdot \frac{d_{i, j}}{\left\|d_{i, j}\right\|}\right)
\end{equation}

After that, these three feature terms are weighted and aggregated as the final constraint term:

\begin{equation}
    W(i,j)=\frac{1}{1+(\gamma_1 d_{i,j}+\gamma_2 \theta_{d_{i,j}}+\gamma_3 \theta_{i,j})},  
\end{equation}
where $\gamma_1, \gamma_2, \gamma_3$ are the weight parameters. 
During training, we supervise the learned attention map with this PPF constraint term

\begin{equation}
L_{attention}=\frac{1}{NN}\sum_{i=1}^{N}\sum_{j=1}^{N} (M(i,j)-W(i,j))^2
\end{equation}



\subsection{Attention Aware Pose Estimation}
Given the extracted features $F_s$ and $F_m$, and the point-wise attention map $M$, we design an attention aware pose estimation mechanism. The key idea is to concatenate the attention map to the feature vectors to guide the correspondence matching and pose estimating process.

Specifically, we follow~\cite{43} to develop two different branches to separately predict the point matches $C_{s\rightarrow m}$ from the scene points to model points and the point matches $C_{m\rightarrow s}$ from the model points to scene points. In each branch, an MLP is deployed to decode the feature vector $F_s$ or $F_m$, then the output features are concatenated with the attention map and fed to an MLP regressor for correspondence prediction. In order to encode more features, we also deploy inter-branch concatenation and short circuits mechanism as shown in Fig.~\ref{fig:network}.
Given predicted scene to model matches $(x_i,n_i)$ and model to scene matches $(x_j,n_j)$,
we directly supervise the predicted correspondence with $L1$ losses
\begin{equation}
    L_s=\frac{1}{N} \sum_i\left(\left\|x_i-\hat{x}_i\right\|+\varepsilon\left\|n_i^s-\hat{n}_i^s\right\|\right)
\end{equation}
\begin{equation}
    L_m=\frac{1}{N} \sum_j\left(\left\|x_j-\widehat{x}_j\right\|+\varepsilon\left\|n_j^m-\hat{n}_j^m\right\|\right)
\end{equation}
where $\varepsilon$ is a hyper-parameter to balance the two terms.

Moreover, we also adopt another regression branch to directly predict the candidate poses $T_d=(R_{d},t_{d})$ from decoded features following~\cite{9} by regressing the 3D translation vector and a normalized 4D quaternion vector, in which the attention map is also concatenated in the way as in the other two branches. The poses are supervised with the ground truth pose with ADD loss for asymmetric objects
\begin{equation}
    L_{d_{i}}=\frac{1}{{K}} \sum_{k}\left\|\left({R_{d_{i}}} {p}_{k}+{t_{d_{i}}}\right)-\left(\hat{{R}}_{d_{i}} {p}_{k}+\hat{{t}}_{d_{i}} \right)\right\|
\end{equation}
or with ADD-S loss for symmetric objects
\begin{equation}
    L_{d_{i}}=\frac{1}{K} \sum_{k} \min_{j \in K}\left\|\left(R_{d_{i}} p_j+t_{d_{i}}\right)-\left(\hat{R}_{d_{i}} p_k+\hat{t}_{d_{i}}\right)\right\|
\end{equation}


We train the network end-to-end with prediction losses, pose losses, and the attention loss together 
\begin{equation}
    L=\varphi_1\frac{1}{N} \sum_i L_{d_{i}}+\varphi_2 L_s+\varphi_3 L_m + \varphi_4 L_{attention}
\end{equation}

Last, we follow~\cite{24} to compute the possible poses $T_s$ and $T_m$ from the predicted point pairs. And due to the complementary nature of the information in these three pose sets, we merge the predicted poses from the three branches to obtain the final pose:
\begin{equation}
    {T}_{final}=average({T}_{d} \cup {T}_{{s}} \cup {T}_{{m}})
\end{equation}

\begin{table*}[h]
\caption{Evaluation results in terms of ADD(-S)($<$0.1d) on LineMOD dataset}\label{tbl1}
\vspace{-4mm}
\begin{center}
\begin{threeparttable}

    \begin{tabular}{c|c|c|c|c|c|c|c|c}
\hline
& DenseFusion \cite{11} & GCCN \cite{8} & REDE \cite{hua2021rede} & G2L-Net \cite{20} & PR-GCN \cite{40} & PVN3D \cite{34} & BiCo-net \cite{43} & Ours \\
\hline
ape & 92.3 & 97.5 & 95.6 & 96.8 & 97.6 & 97.3 & 97.3 & \textbf{98.2} \\
benchvise & 93.2 & 98.5 & 99.4 & 96.1 & 99.2 & 99.7 & 98.8 & \textbf{99.7} \\
camera & 94.4 & 99.7 & 99.6 & 98.2 & 99.4 & 99.6 & 99.6 & \textbf{100.0} \\
can & 93.1 & 99.5 & 99.5 & 98.0 & 98.4 & 99.5 & 99.3 & \textbf{99.8} \\
cat & 96.5 & 98.8 & 99.5 & 99.2 & 98.7 & 99.8 & 100.0 & \textbf{100.0} \\
driller & 87.0 & 96.6 & 99.3 & \textbf{99.8} & 98.8 & 99.3 & 98.9 & 99.3 \\
duck & 92.3 & 98.7 & 97.0 & 97.7 & 98.9 & 98.2 & 98.7 & \textbf{99.0} \\
eggbox$^\ast$ & 99.8 & \textbf{100.0} & \textbf{100.0} & \textbf{100.0} & 99.9 & 99.8 & 99.8 & 99.8 \\
glue$^\ast$ & \textbf{100.0} & \textbf{100.0} & 99.9 & \textbf{100.0} & \textbf{100.0} & \textbf{100.0} & 99.8 & 99.9 \\
holepuncher & 92.1 & 96.7 & 98.6 & 99.0 & 99.4 & \textbf{99.9} & 99.2 & 99.8 \\
iron & 97.0 & 97.1 & 99.3 & 99.3 & 98.5 & 99.7 & \textbf{100.0} & 99.9 \\
lamp & 95.3 & 99.1 & 99.3 & 99.5 & 99.2 & 99.8 & 99.7 & \textbf{99.8} \\
phone & 92.8 & 98.4 & 99.3 & 98.9 & 98.4 & 99.5 & 99.2 & \textbf{99.5} \\
\hline
MEAN & 94.3 & 98.5 & 98.9 & 98.7 & 98.9 & 99.4 & 99.3 & \textbf{99.6} \\
\hline
\end{tabular}
\begin{tablenotes}
    \footnotesize
    \item[*] Objects marked with stars  are symmetrical objects
\end{tablenotes} 
\end{threeparttable}
\end{center}
\vspace{-6mm}
\end{table*}

\section{Experiment and Discussion}

This section presents our experimental setup and implementation details and then reports the evaluation results on several commonly used datasets. We also demonstrate the effectiveness of our proposed components by performing several ablation studies.

\subsection{Datasets}
LineMOD \cite{7} dataset contains a total of 13 objects, and we follow the approach in \cite{9} to segment the training and testing data. Specifically, the dataset contains 13 low-texture objects placed in different cluttered environments, comprising 15783 images. And 1065 real data are randomly selected from the original dataset for testing.

YCB-Video \cite{9} contains 21 shape and texture variations of YCB \cite{42} objects. A subset of 92 RGBD videos of the objects is captured and annotated using 6D poses and instance semantic masks. 

We follow \cite{11} to use the GT mask for training, and divide the dataset into training and testing sets. 16189 frames plus 80,000 synthetic images provided by \cite{9} are selected for training, and another 2949 critical frames from the remaining 12 videos are selected for testing.

Occ-LineMOD \cite{41} dataset is a subset of the LineMOD dataset, containing 8 objects under severe occlusion and 1214 images with multiple severely occluded objects. We use this dataset to test the robustness of pose estimation in challenging situations.



\subsection{Evalution Metrics}
We used ADD \cite{7} and ADD-S \cite{9} evaluation metrics used by most methods to evaluate our model.
ADD is the average Euclidean distance between the model points after transforming the predicted and ground truth poses.
ADD-S is a metric for symmetric objects to calculate the average distance to the nearest point.
In both LineMOD and Occ-LineMOD datasets, we report the accuracy of pose prediction for ADD(-S) $<$ 0.1d. While for the YCB-Video dataset, we report the area under the curve obtained by ADD(-S) by varying the distance threshold and the percentage of all ADD(-S) data less than 2 cm.

\subsection{Experimental results}
\textbf{LineMOD.} 
We evaluate our performance in the LineMOD dataset as shown in Table \ref{tbl1}. 
For a fair comparison, apart from~\cite{34} which predicts masks by their own, all other methods including ours utilize the masks provided by PoseCNN~\cite{9}.
Our method uses only real data for training and outperforms all other methods, with a higher accuracy of more than 0.2\%. 
For small objects in the dataset, it is demanding for other networks to estimate their poses effectively based on a small number of pixel points. Our method tightly links the model point cloud and depth information. Eventually, the prediction robustness of these objects is significantly improved compared to other methods.
The performance of our approach exhibits a slight deficiency when applied to the symmetrical objects. We argue that it is attributed to the existence of multimodal responses in the learned attention maps that represent the correlation between model points and scene points, pertaining to the geometric symmetry of the objects. Consequently, these multimodal responses marginally impact the learning of the matching process.

\begin{figure*}
	\centering	
            \includegraphics[scale=.35]{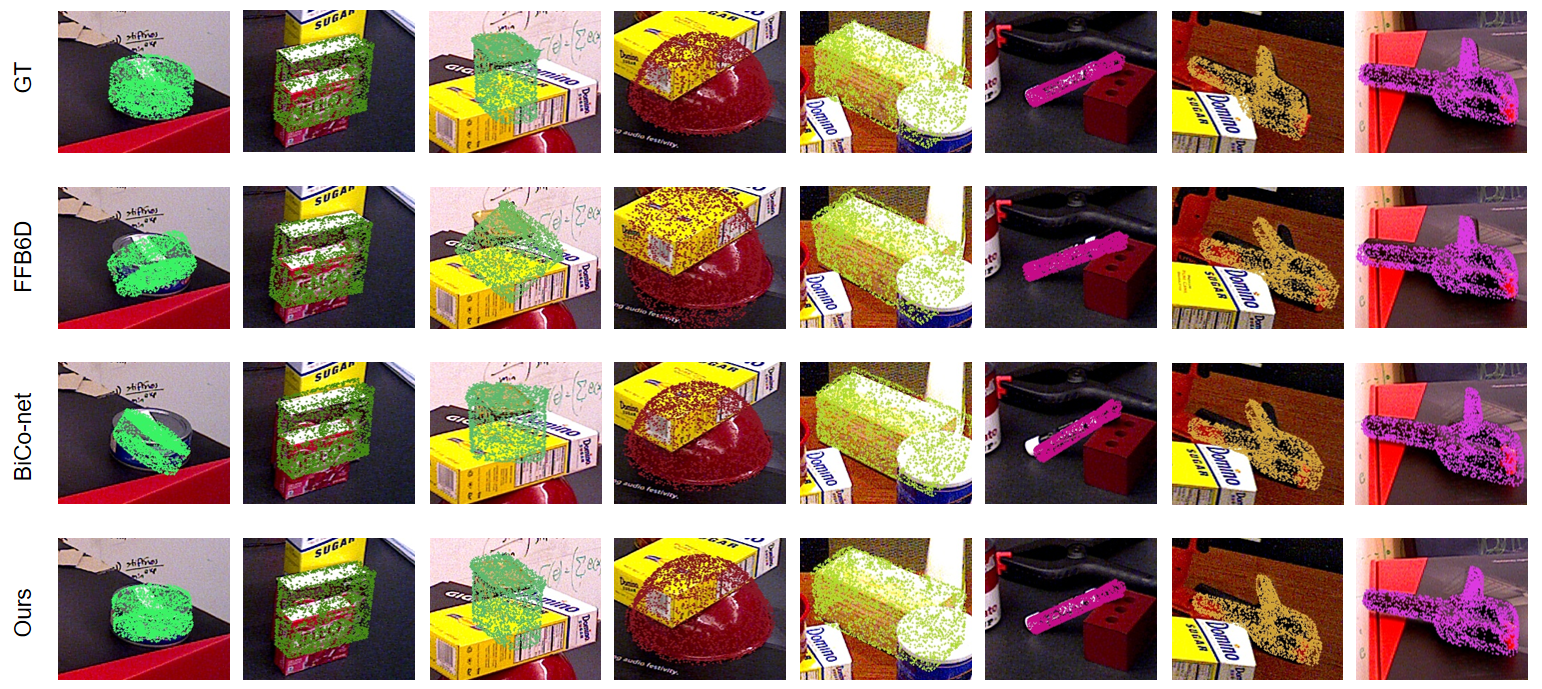}
            \vspace{-4mm}
	\caption{Illustration of the performance of our method compared with other baseline methods on the YCB-Video dataset. Point clouds are projected back to the image after being transformed by the predicted pose. Images are cropped for better visualization.}
	\label{FIG:7}
 \vspace{-4mm}
\end{figure*}

\textbf{YCB-Video.} 
We evaluate our performance with the GT mask and the PVN3D\cite{34} masks respectively for a fair comparison, as shown in Table~\ref{tbl2}.
Our method has an advantage over most of the state-of-the-art methods and achieves 99.1\% on ADD-S ($<$2cm). 
Our accuracy improves on most objects thanks to the point-pair feature constraint. Fig.~\ref{FIG:7} shows the results of the visualization of the predicted poses of some of the objects. It can be seen that our method has improved the robustness of the network in predicting poses to some extent, and is able to accurately calculate the correct poses even with interference such as occlusion.

\begin{table*}[h]
  \caption{Evaluation results in terms of the ADD-S(AUC) and ADD-S($<$2cm) evaluation metrics on YCB-Video dataset}\label{tbl2}
  \vspace{-4mm}
  \begin{center}
  \begin{threeparttable}
    \begin{tabular}{c|c|c|c|c|c|c|c|c|c|c|c|c|c|c|c|c}
    \hline
    \multicolumn{1}{c}{ } & \multicolumn{6}{|c}{with GT mask} &  \multicolumn{8}{|c}{with PVN3D mask} \\
    \hline
    \multicolumn{1}{c}{ } & \multicolumn{2}{|c}{DenseFusion \cite{11}} & \multicolumn{2}{|c}{BiCo-net \cite{43}} & \multicolumn{2}{|c}{Ours} & \multicolumn{2}{|c}{PVN3D \cite{34}} & \multicolumn{2}{|c}{PR-GCN \cite{40}} & \multicolumn{2}{|c}{BiCo-net \cite{43}} & \multicolumn{2}{|c}{Ours} \\
    \hline
    
    \multicolumn{1}{c}{Object} & \multicolumn{1}{|c}{AUC} & \multicolumn{1}{|c}{$<$2cm} & \multicolumn{1}{|c}{AUC} & \multicolumn{1}{|c}{$<$2cm} & \multicolumn{1}{|c}{AUC} & \multicolumn{1}{|c}{$<$2cm} & \multicolumn{1}{|c}{AUC} & \multicolumn{1}{|c}{$<$2cm} & \multicolumn{1}{|c}{AUC} & \multicolumn{1}{|c}{$<$2cm}  & \multicolumn{1}{|c}{AUC} & \multicolumn{1}{|c}{$<$2cm} & \multicolumn{1}{|c}{AUC} & \multicolumn{1}{|c}{$<$2cm}  \\    
    \hline

    \multicolumn{1}{c}{002} & \multicolumn{1}{|c}{96.2} & \multicolumn{1}{|c}{100.0} & \multicolumn{1}{|c}{96.3} & \multicolumn{1}{|c}{100} & \multicolumn{1}{|c}{\textbf{96.9}} & \multicolumn{1}{|c}{\textbf{100}} & \multicolumn{1}{|c}{96.0} & \multicolumn{1}{|c}{100} & \multicolumn{1}{|c}{\textbf{97.1}} & \multicolumn{1}{|c}{100} & \multicolumn{1}{|c}{96.4} & \multicolumn{1}{|c}{100.0} & \multicolumn{1}{|c}{95.8} & \multicolumn{1}{|c}{\textbf{100}} \\
    
    \multicolumn{1}{c}{003} & \multicolumn{1}{|c}{95.3} & \multicolumn{1}{|c}{100} & \multicolumn{1}{|c}{96.5} & \multicolumn{1}{|c}{100} & \multicolumn{1}{|c}{\textbf{96.2}} & \multicolumn{1}{|c}{\textbf{100}} & \multicolumn{1}{|c}{96.1} & \multicolumn{1}{|c}{100} & \multicolumn{1}{|c}{\textbf{97.6}} & \multicolumn{1}{|c}{100} & \multicolumn{1}{|c}{96.1} & \multicolumn{1}{|c}{99.9} & \multicolumn{1}{|c}{96.5} & \multicolumn{1}{|c}{\textbf{100}} \\
    
    \multicolumn{1}{c}{004} & \multicolumn{1}{|c}{97.9} & \multicolumn{1}{|c}{100} & \multicolumn{1}{|c}{97.5} & \multicolumn{1}{|c}{100} & \multicolumn{1}{|c}{\textbf{98}} & \multicolumn{1}{|c}{\textbf{100}} &  \multicolumn{1}{|c}{97.4} & \multicolumn{1}{|c}{100} & \multicolumn{1}{|c}{\textbf{98.3}} & \multicolumn{1}{|c}{100} & \multicolumn{1}{|c}{97.9} & \multicolumn{1}{|c}{100} & \multicolumn{1}{|c}{98.1} & \multicolumn{1}{|c}{\textbf{100}} \\
    
    \multicolumn{1}{c}{005} & \multicolumn{1}{|c}{94.3} & \multicolumn{1}{|c}{96.9} & \multicolumn{1}{|c}{96.4} & \multicolumn{1}{|c}{98.7} & \multicolumn{1}{|c}{\textbf{96.8}} & \multicolumn{1}{|c}{\textbf{98.7}} & \multicolumn{1}{|c}{\textbf{96.2}} & \multicolumn{1}{|c}{98.1} & \multicolumn{1}{|c}{95.3} & \multicolumn{1}{|c}{97.6} & \multicolumn{1}{|c}{95.8} & \multicolumn{1}{|c}{\textbf{98.1}} & \multicolumn{1}{|c}{95.9} & \multicolumn{1}{|c}{98} \\
    
    \multicolumn{1}{c}{006} & \multicolumn{1}{|c}{97.7} & \multicolumn{1}{|c}{100} & \multicolumn{1}{|c}{98.0} & \multicolumn{1}{|c}{100} & \multicolumn{1}{|c}{\textbf{98}} & \multicolumn{1}{|c}{\textbf{100}} & \multicolumn{1}{|c}{97.5} & \multicolumn{1}{|c}{100} & \multicolumn{1}{|c}{97.9} & \multicolumn{1}{|c}{100} & \multicolumn{1}{|c}{97.9} & \multicolumn{1}{|c}{100} & \multicolumn{1}{|c}{\textbf{98.3}} & \multicolumn{1}{|c}{\textbf{100}} \\
    
    \multicolumn{1}{c}{007} & \multicolumn{1}{|c}{\textbf{96.7}} & \multicolumn{1}{|c}{100} & \multicolumn{1}{|c}{95.9} & \multicolumn{1}{|c}{100} & \multicolumn{1}{|c}{96.5} & \multicolumn{1}{|c}{\textbf{100}} & \multicolumn{1}{|c}{96} & \multicolumn{1}{|c}{100} & \multicolumn{1}{|c}{\textbf{97.6}} & \multicolumn{1}{|c}{100} & \multicolumn{1}{|c}{96.2} & \multicolumn{1}{|c}{100} & \multicolumn{1}{|c}{97.1} & \multicolumn{1}{|c}{\textbf{100}} \\
    
    \multicolumn{1}{c}{008} & \multicolumn{1}{|c}{97.3} & \multicolumn{1}{|c}{100} & \multicolumn{1}{|c}{97.7} & \multicolumn{1}{|c}{100} & \multicolumn{1}{|c}{\textbf{97.8}} & \multicolumn{1}{|c}{\textbf{100}} & \multicolumn{1}{|c}{97.1} & \multicolumn{1}{|c}{100} & \multicolumn{1}{|c}{\textbf{98.4}} & \multicolumn{1}{|c}{100} & \multicolumn{1}{|c}{97.3} & \multicolumn{1}{|c}{100} & \multicolumn{1}{|c}{98.1} & \multicolumn{1}{|c}{\textbf{100}} \\
    
    \multicolumn{1}{c}{009} & \multicolumn{1}{|c}{98.4} & \multicolumn{1}{|c}{100} & \multicolumn{1}{|c}{98.3} & \multicolumn{1}{|c}{100} & \multicolumn{1}{|c}{\textbf{98.9}} & \multicolumn{1}{|c}{\textbf{100}} & \multicolumn{1}{|c}{97.7} & \multicolumn{1}{|c}{100} & \multicolumn{1}{|c}{96.2} & \multicolumn{1}{|c}{94.4} & \multicolumn{1}{|c}{\textbf{98.9}} & \multicolumn{1}{|c}{100} & \multicolumn{1}{|c}{98.7} & \multicolumn{1}{|c}{\textbf{100}} \\
    
    \multicolumn{1}{c}{010} & \multicolumn{1}{|c}{90.2} & \multicolumn{1}{|c}{92.3} & \multicolumn{1}{|c}{93.1} & \multicolumn{1}{|c}{\textbf{95.6}} & \multicolumn{1}{|c}{\textbf{93.6}} & \multicolumn{1}{|c}{95.4} & \multicolumn{1}{|c}{93.3} & \multicolumn{1}{|c}{94.6} & \multicolumn{1}{|c}{\textbf{96.6}} & \multicolumn{1}{|c}{\textbf{99.1}} & \multicolumn{1}{|c}{93} & \multicolumn{1}{|c}{94.7} & \multicolumn{1}{|c}{93.4} & \multicolumn{1}{|c}{94.8} \\
    
    \multicolumn{1}{c}{011} & \multicolumn{1}{|c}{96.2} & \multicolumn{1}{|c}{99.7} & \multicolumn{1}{|c}{97.4} & \multicolumn{1}{|c}{100} & \multicolumn{1}{|c}{\textbf{97.7}} & \multicolumn{1}{|c}{\textbf{100}} & \multicolumn{1}{|c}{96.6} & \multicolumn{1}{|c}{100} & \multicolumn{1}{|c}{\textbf{98.5}} & \multicolumn{1}{|c}{100} & \multicolumn{1}{|c}{97.4} & \multicolumn{1}{|c}{100} & \multicolumn{1}{|c}{97.6} & \multicolumn{1}{|c}{\textbf{100}} \\
    
    \multicolumn{1}{c}{019} & \multicolumn{1}{|c}{97.5} & \multicolumn{1}{|c}{100} & \multicolumn{1}{|c}{97.0} & \multicolumn{1}{|c}{100} & \multicolumn{1}{|c}{\textbf{97.5}} & \multicolumn{1}{|c}{\textbf{100}} & \multicolumn{1}{|c}{97.4} & \multicolumn{1}{|c}{100} & \multicolumn{1}{|c}{\textbf{98.1}} & \multicolumn{1}{|c}{100} & \multicolumn{1}{|c}{97.5} & \multicolumn{1}{|c}{100} & \multicolumn{1}{|c}{97.6} & \multicolumn{1}{|c}{\textbf{100}} \\
    
    \multicolumn{1}{c}{021} & \multicolumn{1}{|c}{96.4} & \multicolumn{1}{|c}{100} & \multicolumn{1}{|c}{97.0} & \multicolumn{1}{|c}{100} & \multicolumn{1}{|c}{\textbf{97.1}} & \multicolumn{1}{|c}{\textbf{100}} & \multicolumn{1}{|c}{96} & \multicolumn{1}{|c}{100} & \multicolumn{1}{|c}{\textbf{97.9}} & \multicolumn{1}{|c}{100} & \multicolumn{1}{|c}{96.4} & \multicolumn{1}{|c}{100} & \multicolumn{1}{|c}{96.8} & \multicolumn{1}{|c}{\textbf{100}} \\
    
    \multicolumn{1}{c}{024$^\ast$} & \multicolumn{1}{|c}{88.9} & \multicolumn{1}{|c}{87.4} & \multicolumn{1}{|c}{96.5} & \multicolumn{1}{|c}{100} & \multicolumn{1}{|c}{\textbf{97.0}} & \multicolumn{1}{|c}{\textbf{100}} & \multicolumn{1}{|c}{90.2} & \multicolumn{1}{|c}{80.5} & \multicolumn{1}{|c}{90.3} & \multicolumn{1}{|c}{96.6} & \multicolumn{1}{|c}{\textbf{96.5}} & \multicolumn{1}{|c}{\textbf{100}} & \multicolumn{1}{|c}{96} & \multicolumn{1}{|c}{99.3} \\
    
    \multicolumn{1}{c}{025} & \multicolumn{1}{|c}{97.0} & \multicolumn{1}{|c}{100} & \multicolumn{1}{|c}{96.5} & \multicolumn{1}{|c}{100} & \multicolumn{1}{|c}{\textbf{97.3}} & \multicolumn{1}{|c}{\textbf{100}} & \multicolumn{1}{|c}{97.6} & \multicolumn{1}{|c}{100} & \multicolumn{1}{|c}{\textbf{98.1}} & \multicolumn{1}{|c}{100} & \multicolumn{1}{|c}{97.2} & \multicolumn{1}{|c}{100} & \multicolumn{1}{|c}{97.4} & \multicolumn{1}{|c}{\textbf{100}} \\
    
    \multicolumn{1}{c}{035} & \multicolumn{1}{|c}{97.1} & \multicolumn{1}{|c}{100} & \multicolumn{1}{|c}{96.8} & \multicolumn{1}{|c}{100} & \multicolumn{1}{|c}{\textbf{97.1}} & \multicolumn{1}{|c}{\textbf{100}} & \multicolumn{1}{|c}{96.7} & \multicolumn{1}{|c}{100} & \multicolumn{1}{|c}{\textbf{98.1}} & \multicolumn{1}{|c}{100} & \multicolumn{1}{|c}{96.9} & \multicolumn{1}{|c}{100} & \multicolumn{1}{|c}{97.3} & \multicolumn{1}{|c}{\textbf{100}} \\
    
    \multicolumn{1}{c}{036$^\ast$} & \multicolumn{1}{|c}{94.1} & \multicolumn{1}{|c}{100} & \multicolumn{1}{|c}{95.2} & \multicolumn{1}{|c}{100} & \multicolumn{1}{|c}{\textbf{95.2}} & \multicolumn{1}{|c}{\textbf{100}} & \multicolumn{1}{|c}{90.4} & \multicolumn{1}{|c}{93.8} & \multicolumn{1}{|c}{\textbf{96}} & \multicolumn{1}{|c}{\textbf{100}} & \multicolumn{1}{|c}{91.5} & \multicolumn{1}{|c}{89.7} & \multicolumn{1}{|c}{94.1} & \multicolumn{1}{|c}{98.8} \\
    
    \multicolumn{1}{c}{037} & \multicolumn{1}{|c}{93.2} & \multicolumn{1}{|c}{100} & \multicolumn{1}{|c}{\textbf{95}} & \multicolumn{1}{|c}{100} & \multicolumn{1}{|c}{94.5} & \multicolumn{1}{|c}{\textbf{100}} & \multicolumn{1}{|c}{\textbf{96.7}} & \multicolumn{1}{|c}{100} & \multicolumn{1}{|c}{\textbf{96.7}} & \multicolumn{1}{|c}{100} & \multicolumn{1}{|c}{90.8} & \multicolumn{1}{|c}{98.9} & \multicolumn{1}{|c}{93.5} & \multicolumn{1}{|c}{\textbf{100}} \\
    
    \multicolumn{1}{c}{040} & \multicolumn{1}{|c}{97.5} & \multicolumn{1}{|c}{100} & \multicolumn{1}{|c}{\textbf{97.3}} & \multicolumn{1}{|c}{100} & \multicolumn{1}{|c}{97.2} & \multicolumn{1}{|c}{\textbf{100}} & \multicolumn{1}{|c}{96.7} & \multicolumn{1}{|c}{99.8} & \multicolumn{1}{|c}{97.9} & \multicolumn{1}{|c}{100} & \multicolumn{1}{|c}{96.8} & \multicolumn{1}{|c}{100} & \multicolumn{1}{|c}{\textbf{98}} & \multicolumn{1}{|c}{\textbf{100}} \\

    \multicolumn{1}{c}{051$^\ast$} & \multicolumn{1}{|c}{89.7} & \multicolumn{1}{|c}{98.0} & \multicolumn{1}{|c}{95.9} & \multicolumn{1}{|c}{100} & \multicolumn{1}{|c}{\textbf{95.9}} & \multicolumn{1}{|c}{\textbf{100}} & \multicolumn{1}{|c}{93.6} & \multicolumn{1}{|c}{93.6} & \multicolumn{1}{|c}{87.5} & \multicolumn{1}{|c}{93.3} & \multicolumn{1}{|c}{94.4} & \multicolumn{1}{|c}{98.5} & \multicolumn{1}{|c}{92} & \multicolumn{1}{|c}{\textbf{98.5}} \\
    
    \multicolumn{1}{c}{052$^\ast$} & \multicolumn{1}{|c}{77.4} & \multicolumn{1}{|c}{80.5} & \multicolumn{1}{|c}{95.1} & \multicolumn{1}{|c}{99.9} & \multicolumn{1}{|c}{\textbf{95.6}} & \multicolumn{1}{|c}{\textbf{100}} & \multicolumn{1}{|c}{88.4} & \multicolumn{1}{|c}{83.6} & \multicolumn{1}{|c}{79.7} & \multicolumn{1}{|c}{84.6} & \multicolumn{1}{|c}{88.4} & \multicolumn{1}{|c}{91.2} & \multicolumn{1}{|c}{86.9} & \multicolumn{1}{|c}{\textbf{91.5}} \\
    
    \multicolumn{1}{c}{061$^\ast$} & \multicolumn{1}{|c}{91.5} & \multicolumn{1}{|c}{100} & \multicolumn{1}{|c}{96.8} & \multicolumn{1}{|c}{100} & \multicolumn{1}{|c}{\textbf{97.3}} & \multicolumn{1}{|c}{\textbf{100}} & \multicolumn{1}{|c}{96.8} & \multicolumn{1}{|c}{100} & \multicolumn{1}{|c}{\textbf{97.8}} & \multicolumn{1}{|c}{100} & \multicolumn{1}{|c}{97.2} & \multicolumn{1}{|c}{100} & \multicolumn{1}{|c}{96.9} & \multicolumn{1}{|c}{\textbf{100}} \\

   \hline
    \multicolumn{1}{c}{ALL} & \multicolumn{1}{|c}{94.2} & \multicolumn{1}{|c}{97.8} & \multicolumn{1}{|c}{96.4} & \multicolumn{1}{|c}{99.6} & \multicolumn{1}{|c}{\textbf{96.7}} & \multicolumn{1}{|c}{\textbf{96.6}} & \multicolumn{1}{|c}{95.5} & \multicolumn{1}{|c}{97.6} & \multicolumn{1}{|c}{95.8} & \multicolumn{1}{|c}{98.5} & \multicolumn{1}{|c}{95.8} & \multicolumn{1}{|c}{98.8} & \multicolumn{1}{|c}{\textbf{95.8}} & \multicolumn{1}{|c}{\textbf{99.1}} \\
    \hline
    \end{tabular}
    \begin{tablenotes}
    \footnotesize
    \item[*] Objects marked with stars are symmetrical objects
    \end{tablenotes}
  \end{threeparttable}
  \end{center}
  \vspace{-6.5mm}
\end{table*}

\textbf{Occ-LineMOD.} For the most challenging dataset, the final results are shown in Table \ref{tbl3}. 
It can be seen that our method shows a significant improvement in ADD-S$<$0.1d compared to other methods, with an average prediction accuracy of 74.4\%.
Notably, our method demonstrates improved performance on symmetrical objects in more occluded situations. We attribute this observation to the reduction in potential correlations between point pairs brought about by surface occlusions. This reduction in the likelihood of multimodal responses subsequently enhances the precision of the matching process.
However, since the structure of PR-GCN \cite{40} based on graph convolutional network can make full use of the geometric information and topology of objects in images, they also show better performance at handling topological information of objects.

\begin{table*}[h]
\caption{Evaluation results in terms of ADD(-S)($<$0.1d) on Occ-LineMOD dataset}\label{tbl3}
\vspace{-4mm}
\begin{center}
\begin{threeparttable}
\begin{tabular}{c|c|c|c|c|c|c}
\hline
& PVNet \cite{17} & REDE \cite{hua2021rede} & FFB6D \cite{12} & PR-GCN \cite{40} & BiCo-net \cite{43} & Ours \\
\hline
ape & 15.8 & 53.1 & 47.2 & 40.2 & 55.6 & \textbf{58.3} \\
can & 63.3 & 88.5 & 85.2 & 76.2 & 83.2 & \textbf{88.5} \\
cat & 16.7 & 35.9 & 45.7 & \textbf{57.0} & 47.3 & 51.6 \\
driller & 65.7 & 77.8 & 81.4 & \textbf{82.3} & 69.9 & 77.8 \\
duck & 25.2 & 46.2 & 53.9 & 30.0 & 58.3 & \textbf{64.8} \\
eggbox$^\ast$ & 50.2 & 71.8 & 70.2 & 68.2 & 78.1 & \textbf{81.3} \\
glue$^\ast$ & 49.6 & 75.0 & 60.1 & 67.0 & 76.9 & \textbf{79.0} \\
holepuncher & 39.7 & 75.5 & 85.9 & \textbf{97.2} & 87.2 & 93.6 \\
\hline
Mean & 40.8 & 65.4 & 66.2 & 65.0 & 69.5 & \textbf{74.4} \\
\hline
\end{tabular}
\begin{tablenotes}
    \footnotesize
    \item[*] Objects marked with stars are symmetrical objects
\end{tablenotes}
\end{threeparttable}
\end{center}
\vspace{-5mm}
\end{table*}

\subsection{Ablation Studies}
\textbf{Effect of attention aware pose estimation.}
To verify the impact of the point-wise attention map module on the robustness of the predicted poses, we conduct ablation studies on the LineMOD and Occ-LineMOD datasets. 
As shown in Table \ref{tbl4}, we find that concatenating the attention map to the 3 branches makes them aware of the correlation between model points and scene observations, which significantly improves the subsequent pose prediction results.



\begin{table}[h]
\caption{Effect of attention aware pose estimation.}\label{tbl4}
\vspace{-4mm}
\begin{center}
\begin{tabular}{c|c|c|c}
\hline
\multicolumn{1}{p{6.8em}|}{Attention for direct regression} & \multicolumn{1}{p{6.8em}|}{Attention for match prediction} & \multicolumn{1}{p{4em}|}{LineMOD  ADD(-S)} & \multicolumn{1}{p{6.5em}}{Occ-LineMOD ADD(-S)}\\
\hline
  &  & 99.3 & 69.5 \\
 & \checkmark & 99.4 & 72.4 \\
\checkmark &  & 99.5 & 73.7 \\
\checkmark & \checkmark & 99.6 & 74.4 \\
\hline
\end{tabular}
\end{center}
\vspace{-5mm}
\end{table}


\textbf{Effect of supervising the attention map with PPF features.}
To verify the effect of the point-pair feature weights in supervising the point-wise attention map, we recombine these three features and conduct ablation studies on the LineMOD and Occ-LineMOD datasets. As shown in Table \ref{tbl5}, if the PPF constraints term only uses the point distance vector feature and the angle of the normal feature, it ends up with 73.3\% of the final results, which is a 3.8\% improvement compared to the original network without using constrained weights. However, we add the angle between the normal and the distance vector as another feature constraint to jointly guide the global attention map, and the final result is improved by 1.1\%. 


\begin{table}[h]
\caption{Effect of PPF weight constraint terms}\label{tbl5}
\vspace{-4mm}
\begin{center}
\begin{tabular}{c|c|c|c|c}
\hline
$d$ & $\theta_{d, N}$ & $\theta_{N}$ & \multicolumn{1}{p{4em}|}{LineMOD  ADD(-S)} & \multicolumn{1}{p{6.5em}}{Occ-LineMOD ADD(-S)}\\
\hline
\checkmark &  &  & 99.5 & 73.4\\
 & \checkmark &  & 99.4 & 73.5 \\
 &  & \checkmark & 99.4 & 73.9 \\
\checkmark &  & \checkmark & 99.4 & 73.3 \\
\checkmark & \checkmark &  & 99.5 & 73.5 \\
 & \checkmark & \checkmark & 99.4 & 74.0 \\
\checkmark & \checkmark & \checkmark & 99.6 & 74.4\\
\hline
\end{tabular}
\end{center}
\vspace{-5.5mm}
\end{table}

\begin{figure}
	\centering	
            \includegraphics[scale=.22]{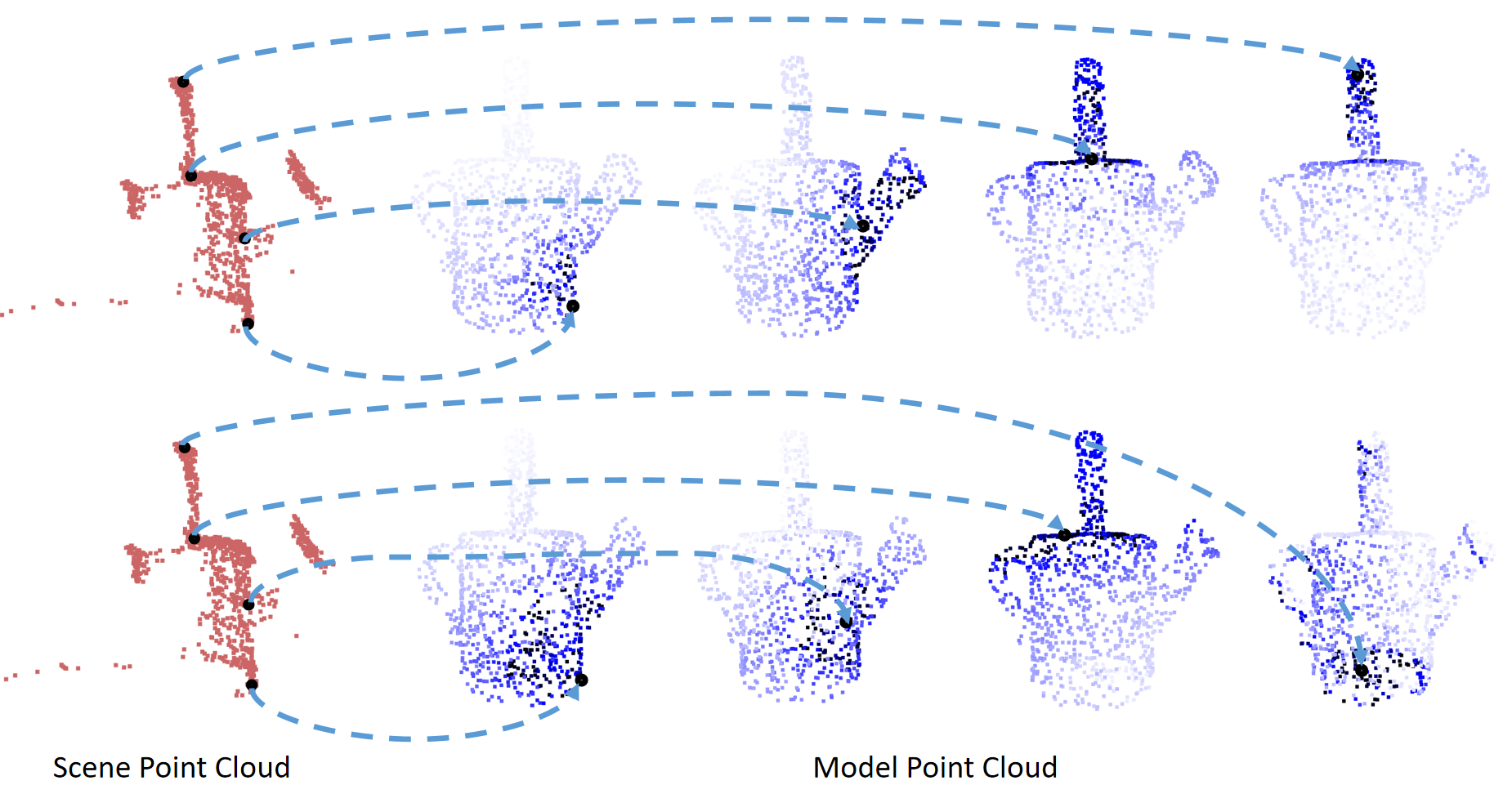}
            \vspace{-4mm}
    \caption{Visualization of attention maps learned by our attention module (first row) and by GCCN method (second row). We select a point in the scene point cloud, and the corresponding attention map on the model point cloud is projected into a 2D image plane for visualization. The dotted lines connect the scene point and the corresponding points with the largest attention values.}
	\label{FIG:8}
\vspace{-6.5mm} 
\end{figure}

\textbf{Effect of point-wise attention mechanism compared with GCCN.}
In order to validate the effectiveness of our proposed point-wise attention mechanism compared with GCCN~\cite{8}, we conduct a series of ablation experiments on all the three datasets. As shown in Table~\ref{tbl6}, we first replace the pseudo-siamese network (PSN) to the feature extraction networks in GCCN. The experimental results show a significant decrease in the pose prediction performance compared to using PSN. Then, we apply the improved PPF weight constraint terms and achieve more improvements. Finally, as shown in Fig.~\ref{FIG:8}, we visually compare the results by taking the point clouds from different viewpoints of the "can" object in the Occ-LineMOD dataset. It can be observed that our point-wise attention map can better model the weight distribution that reflects the correlation between the model and the scene, leading to higher matching precision.


\begin{table}[h]
\caption{Effect of pseudo-siamese network (PSN)}\label{tbl6}
\vspace{-4mm}
\begin{center}
\begin{tabular}{c|c|c|c|c}
\hline
\multicolumn{1}{p{2.1875em}|}{PSN} & \multicolumn{1}{p{4.1875em}|}{PPF constraints} & \multicolumn{1}{p{4.1875em}|}{LineMOD (ADD-S)} & \multicolumn{1}{p{6.2875em}|}{Occ-LineMOD (ADD-S)} & \multicolumn{1}{p{5.1875em}}{YCB-Video ($<$2cm)}\\
\hline
 &  & 99.3 & 73.3 & 98.8\\
 & \checkmark & 99.4 & 73.7 & 98.9 \\
\checkmark & \checkmark & 99.6 & 74.4 & 99.1 \\
\hline
\end{tabular}
\end{center}
\vspace{-6.5mm}
\end{table}

\section{Conclusion}
In this paper, we propose a bidirectional correspondence prediction network with point-wise attention aware mechanism to utilize a CAD model for stable 6D object pose estimation. Also, we introduce 
pseudo-siamese network to discover the similarities between model points and scene points, obtaining robust attention correlations.
Experiments display that our method shows advantages over current state-of-the-art methods, and the accuracy and robustness of our prediction results are improved for 6D pose estimation. 
However, our performance still relies on the quality of the mask to a certain extent, which we have not addressed in this paper. We will consider this issue and try to implement associative segmentation and pose estimation in future works.










\bibliographystyle{unsrt}

\bibliography{root}

\begin{thebibliography}{10}

\bibitem{1}
Eric Marchand, Hideaki Uchiyama, and Fabien Spindler.
\newblock Pose estimation for augmented reality: a hands-on survey.
\newblock {\em IEEE transactions on visualization and computer graphics}, 22(12):2633--2651, 2015.

\bibitem{2}
Xiaozhi Chen, Huimin Ma, Ji~Wan, Bo~Li, and Tian Xia.
\newblock Multi-view 3d object detection network for autonomous driving.
\newblock In {\em Proceedings of the IEEE conference on Computer Vision and Pattern Recognition}, pages 1907--1915, 2017.

\bibitem{4}
Menglong Zhu, Konstantinos~G Derpanis, Yinfei Yang, Samarth Brahmbhatt, Mabel Zhang, Cody Phillips, Matthieu Lecce, and Kostas Daniilidis.
\newblock Single image 3d object detection and pose estimation for grasping.
\newblock In {\em 2014 IEEE International Conference on Robotics and Automation (ICRA)}, pages 3936--3943. IEEE, 2014.

\bibitem{wang2019normalized}
He~Wang, Srinath Sridhar, Jingwei Huang, Julien Valentin, Shuran Song, and Leonidas~J Guibas.
\newblock Normalized object coordinate space for category-level 6d object pose and size estimation.
\newblock In {\em Proceedings of the IEEE/CVF Conference on Computer Vision and Pattern Recognition}, pages 2642--2651, 2019.

\bibitem{chen2020category}
Xu~Chen, Zijian Dong, Jie Song, Andreas Geiger, and Otmar Hilliges.
\newblock Category level object pose estimation via neural analysis-by-synthesis.
\newblock In {\em Computer Vision--ECCV 2020: 16th European Conference, Glasgow, UK, August 23--28, 2020, Proceedings, Part XXVI 16}, pages 139--156. Springer, 2020.

\bibitem{7}
Stefan Hinterstoisser, Stefan Holzer, Cedric Cagniart, Slobodan Ilic, Kurt Konolige, Nassir Navab, and Vincent Lepetit.
\newblock Multimodal templates for real-time detection of texture-less objects in heavily cluttered scenes.
\newblock In {\em 2011 international conference on computer vision}, pages 858--865. IEEE, 2011.

\bibitem{li2018deepim}
Yi~Li, Gu~Wang, Xiangyang Ji, Yu~Xiang, and Dieter Fox.
\newblock Deepim: Deep iterative matching for 6d pose estimation.
\newblock In {\em Proceedings of the European Conference on Computer Vision (ECCV)}, pages 683--698, 2018.

\bibitem{34}
Yisheng He, Yimeng Sun, Jiashuo Huang, Xu~Liu, Chuyang Fan, and Baoquan Li.
\newblock Pvn3d: A deep point-wise 3d keypoints voting network for 6dof pose estimation.
\newblock In {\em Proceedings of the IEEE/CVF Conference on Computer Vision and Pattern Recognition}, pages 5499--5508, 2020.

\bibitem{17}
Sida Peng, Yuan Liu, Qixing Huang, Xiaowei Zhou, and Hujun Bao.
\newblock Pvnet: Pixel-wise voting network for 6dof pose estimation.
\newblock In {\em IEEE Conference on Computer Vision and Pattern Recognition (CVPR)}, pages 4561--4570, 2019.

\bibitem{43}
Zelin Xu, Yichen Zhang, Ke~Chen, and Kui Jia.
\newblock Bico-net: Regress globally, match locally for robust 6d pose estimation.
\newblock {\em arXiv preprint arXiv:2205.03536}, 2022.

\bibitem{8}
Yongming Wen, Yiquan Fang, Junhao Cai, Kimwa Tung, and Hui Cheng.
\newblock Gccn: Geometric constraint co-attention network for 6d object pose estimation.
\newblock In {\em Proceedings of the 29th ACM International Conference on Multimedia}, pages 2671--2679, 2021.

\bibitem{24}
Bertram Drost, Markus Ulrich, Nassir Navab, and Slobodan Ilic.
\newblock Model globally, match locally: Efficient and robust 3d object recognition.
\newblock In {\em 2010 IEEE Computer Society Conference on Computer Vision and Pattern Recognition}, pages 998--1005, 2010.

\bibitem{gordo2016deep}
Albert Gordo, Jon Almaz{\'a}n, Jerome Revaud, and Diane Larlus.
\newblock Deep image retrieval: Learning global representations for image search.
\newblock In {\em Computer Vision--ECCV 2016: 14th European Conference, Amsterdam, The Netherlands, October 11-14, 2016, Proceedings, Part VI 14}, pages 241--257. Springer, 2016.

\bibitem{gu2010discriminative}
Chunhui Gu and Xiaofeng Ren.
\newblock Discriminative mixture-of-templates for viewpoint classification.
\newblock In {\em Computer Vision--ECCV 2010: 11th European Conference on Computer Vision, Heraklion, Crete, Greece, September 5-11, 2010, Proceedings, Part V 11}, pages 408--421. Springer, 2010.

\bibitem{hinterstoisser2013model}
Stefan Hinterstoisser, Vincent Lepetit, Slobodan Ilic, Stefan Holzer, Gary Bradski, Kurt Konolige, and Nassir Navab.
\newblock Model based training, detection and pose estimation of texture-less 3d objects in heavily cluttered scenes.
\newblock In {\em Computer Vision--ACCV 2012: 11th Asian Conference on Computer Vision, Daejeon, Korea, November 5-9, 2012, Revised Selected Papers, Part I 11}, pages 548--562. Springer, 2013.

\bibitem{hodavn2015detection}
Tom{\'a}{\v{s}} Hoda{\v{n}}, Xenophon Zabulis, Manolis Lourakis, {\v{S}}t{\v{e}}p{\'a}n Obdr{\v{z}}{\'a}lek, and Ji{\v{r}}{\'\i} Matas.
\newblock Detection and fine 3d pose estimation of texture-less objects in rgb-d images.
\newblock In {\em 2015 IEEE/RSJ International Conference on Intelligent Robots and Systems (IROS)}, pages 4421--4428. IEEE, 2015.

\bibitem{manhardt2018deep}
Fabian Manhardt, Wadim Kehl, Nassir Navab, and Federico Tombari.
\newblock Deep model-based 6d pose refinement in rgb.
\newblock In {\em Proceedings of the European Conference on Computer Vision (ECCV)}, pages 800--815, 2018.

\bibitem{trabelsi2021pose}
Ameni Trabelsi, Mohamed Chaabane, Nathaniel Blanchard, and Ross Beveridge.
\newblock A pose proposal and refinement network for better 6d object pose estimation.
\newblock In {\em Proceedings of the IEEE/CVF Winter Conference on Applications of Computer Vision}, pages 2382--2391, 2021.

\bibitem{yen2021inerf}
Lin Yen-Chen, Pete Florence, Jonathan~T Barron, Alberto Rodriguez, Phillip Isola, and Tsung-Yi Lin.
\newblock inerf: Inverting neural radiance fields for pose estimation.
\newblock In {\em 2021 IEEE/RSJ International Conference on Intelligent Robots and Systems (IROS)}, pages 1323--1330. IEEE, 2021.

\bibitem{10}
Sergey Zakharov, Ivan Shugurov, and Slobodan Ilic.
\newblock Dpod: 6d pose object detector and refiner.
\newblock In {\em Proceedings of the IEEE/CVF international conference on computer vision}, pages 1941--1950, 2019.

\bibitem{wang2021nemo}
Angtian Wang, Adam Kortylewski, and Alan Yuille.
\newblock Nemo: Neural mesh models of contrastive features for robust 3d pose estimation.
\newblock {\em arXiv preprint arXiv:2101.12378}, 2021.

\bibitem{iwase2021repose}
Shun Iwase, Xingyu Liu, Rawal Khirodkar, Rio Yokota, and Kris~M Kitani.
\newblock Repose: Fast 6d object pose refinement via deep texture rendering.
\newblock In {\em Proceedings of the IEEE/CVF International Conference on Computer Vision}, pages 3303--3312, 2021.

\bibitem{tracknet}
Bowen Wen, Chaitanya Mitash, Baozhang Ren, and Kostas~E Bekris.
\newblock se(3)-tracknet: Data-driven 6d pose tracking by calibrating image residuals in synthetic domains.
\newblock In {\em 2020 IEEE/RSJ International Conference on Intelligent Robots and Systems (IROS)}, pages 10367--10373. IEEE, 2020.

\bibitem{rad2017bb8}
Mahdi Rad and Vincent Lepetit.
\newblock Bb8: A scalable, accurate, robust to partial occlusion method for predicting the 3d poses of challenging objects without using depth.
\newblock In {\em Proceedings of the IEEE international conference on computer vision}, pages 3828--3836, 2017.

\bibitem{tekin2018real}
Bugra Tekin, Sudipta~N Sinha, and Pascal Fua.
\newblock Real-time seamless single shot 6d object pose prediction.
\newblock In {\em Proceedings of the IEEE Conference on Computer Vision and Pattern Recognition}, pages 292--301, 2018.

\bibitem{luo20203d}
Qianhui Luo, Huifang Ma, Li~Tang, Yue Wang, and Rong Xiong.
\newblock 3d-ssd: Learning hierarchical features from rgb-d images for amodal 3d object detection.
\newblock {\em Neurocomputing}, 378:364--374, 2020.

\bibitem{mitash2018robust}
Chaitanya Mitash, Abdeslam Boularias, and Kostas Bekris.
\newblock Robust 6d object pose estimation with stochastic congruent sets.
\newblock {\em arXiv preprint arXiv:1805.06324}, 2018.

\bibitem{wen2020robust}
Bowen Wen, Chaitanya Mitash, Sruthi Soorian, Andrew Kimmel, Avishai Sintov, and Kostas~E Bekris.
\newblock Robust, occlusion-aware pose estimation for objects grasped by adaptive hands.
\newblock In {\em 2020 IEEE International Conference on Robotics and Automation (ICRA)}, pages 6210--6217. IEEE, 2020.

\bibitem{brachmann2019dop}
Eric Brachmann, Carsten Rother, Jens Konrad, and Ben Glocker.
\newblock 6dof object pose estimation via differentiable proxy voting loss.
\newblock In {\em Proceedings of the IEEE International Conference on Computer Vision}, pages 7509--7518, 2019.

\bibitem{chen2022epro}
Hansheng Chen, Pichao Wang, Fan Wang, Wei Tian, Lu~Xiong, and Hao Li.
\newblock Epro-pnp: Generalized end-to-end probabilistic perspective-n-points for monocular object pose estimation.
\newblock In {\em Proceedings of the IEEE/CVF Conference on Computer Vision and Pattern Recognition}, pages 2781--2790, 2022.

\bibitem{hua2021rede}
Weitong Hua, Zhongxiang Zhou, Jun Wu, Huang Huang, Yue Wang, and Rong Xiong.
\newblock Rede: End-to-end object 6d pose robust estimation using differentiable outliers elimination.
\newblock {\em IEEE Robotics and Automation Letters}, 6(2):2886--2893, 2021.

\bibitem{moban}
Chenrui Wu, Long Chen, Shenglong Wang, Han Yang, and Junjie Jiang.
\newblock Geometric-aware dense matching network for 6d pose estimation of objects from rgb-d images.
\newblock {\em Pattern Recognition}, page 109293, 2023.

\bibitem{wang2021gdr}
Gu~Wang, Fabian Manhardt, Federico Tombari, and Xiangyang Ji.
\newblock Gdr-net: Geometry-guided direct regression network for monocular 6d object pose estimation.
\newblock In {\em Proceedings of the IEEE/CVF Conference on Computer Vision and Pattern Recognition}, pages 16611--16621, 2021.

\bibitem{di2021so}
Yan Di, Fabian Manhardt, Gu~Wang, Xiangyang Ji, Nassir Navab, and Federico Tombari.
\newblock So-pose: Exploiting self-occlusion for direct 6d pose estimation.
\newblock In {\em Proceedings of the IEEE/CVF International Conference on Computer Vision}, pages 12396--12405, 2021.

\bibitem{hu2020single}
Yinlin Hu, Pascal Fua, Wei Wang, and Mathieu Salzmann.
\newblock Single-stage 6d object pose estimation.
\newblock In {\em Proceedings of the IEEE/CVF conference on computer vision and pattern recognition}, pages 2930--2939, 2020.

\bibitem{11}
Chen Wang, Danfei Xu, Yuke Zhu, Roberto Mart{\'\i}n-Mart{\'\i}n, Cewu Lu, Li~Fei-Fei, and Silvio Savarese.
\newblock Densefusion: 6d object pose estimation by iterative dense fusion.
\newblock In {\em Proceedings of the IEEE/CVF Conference on Computer Vision and Pattern Recognition}, pages 3343--3352, 2019.

\bibitem{12}
Yisheng He, Zhe Wu, Jun Wang, Ye~Yuan, Zixiang Dong, and Wei Liu.
\newblock Ffb6d: A full flow bidirectional fusion network for 6d pose estimation.
\newblock In {\em Proceedings of the IEEE/CVF Conference on Computer Vision and Pattern Recognition}, pages 5246--5255, 2021.

\bibitem{36}
Zhou Liu, Hongwei Mao, Chuangyu Wu, Yiming Sun, Rongrong Ji, and Zhiquan Su.
\newblock A convnet for the 2020s.
\newblock In {\em Proceedings of the IEEE/CVF Conference on Computer Vision and Pattern Recognition}, pages 11976--11986, 2022.

\bibitem{39}
Charles~R Qi, Hao Su, Kaichun Mo, and Leonidas~J Guibas.
\newblock Pointnet: Deep learning on point sets for 3d classification and segmentation.
\newblock In {\em IEEE Conference on Computer Vision and Pattern Recognition (CVPR)}, pages 652--660, 2017.

\bibitem{9}
Yu~Xiang, Tanner Schmidt, Venkatraman Narayanan, and Dieter Fox.
\newblock Posecnn: A convolutional neural network for 6d object pose estimation in cluttered scenes.
\newblock {\em arXiv preprint arXiv:1711.00199}, 2017.

\bibitem{20}
Wei Chen, Xiao Jia, Hyung~Jin Chang, and et~al.
\newblock G2l-net: Global to local network for real-time 6d pose estimation with embedding vector features.
\newblock In {\em Proceedings of the IEEE/CVF conference on computer vision and pattern recognition}, pages 4233--4242, 2020.

\bibitem{40}
Guangliang Zhou, Hao Wang, Qijun Chen, Yi~Yan, and Deming Wang.
\newblock Pr-gcn: A deep graph convolutional network with point refinement for 6d pose estimation.
\newblock In {\em Proceedings of the IEEE/CVF International Conference on Computer Vision}, pages 2793--2802, 2021.

\bibitem{42}
Berk Calli, Arpit Singh, Ariel Walsman, Siddhartha Srinivasa, Pieter Abbeel, and Aaron~M Dollar.
\newblock The ycb object and model set: Towards common benchmarks for manipulation research.
\newblock In {\em 2015 International Conference on Advanced Robotics (ICAR)}, pages 510--517. IEEE, 2015.

\bibitem{41}
Eric Brachmann, Alexander Krull, Frank Michel, Stefan Gumhold, Jamie Shotton, and Carsten Rother.
\newblock Learning 6d object pose estimation using 3d object coordinates.
\newblock In {\em European conference on computer vision}, pages 536--551. Springer, 2014.

\bibitem{rcvpose}
Yangzheng Wu, Mohsen Zand, Ali Etemad, and Michael Greenspan.
\newblock Vote from the center: 6 dof pose estimation in rgb-d images by radial keypoint voting.
\newblock In {\em European Conference on Computer Vision}, pages 335--352. Springer, 2022.

\bibitem{checkerpose}
Ruyi Lian and Haibin Ling.
\newblock Checkerpose: Progressive dense keypoint localization for object pose estimation with graph neural network.
\newblock {\em arXiv preprint arXiv:2303.16874}, 2023.

\bibitem{zebrapose}
Yongzhi Su, Mahdi Saleh, Torben Fetzer, Jason Rambach, Nassir Navab, Benjamin Busam, Didier Stricker, and Federico Tombari.
\newblock Zebrapose: Coarse to fine surface encoding for 6dof object pose estimation.
\newblock In {\em Proceedings of the IEEE/CVF Conference on Computer Vision and Pattern Recognition}, pages 6738--6748, 2022.

\bibitem{swintf}
Zhujun Li and Ioannis Stamos.
\newblock Depth-based 6dof object pose estimation using swin transformer.
\newblock {\em arXiv preprint arXiv:2303.02133}, 2023.

\end{thebibliography}

\begin{appendices}
\renewcommand{\thesection}{\arabic{section}}%

\section{Details of our method}
\subsection{Implementation Details}
\textbf{Network Architecture.} This experiment runs on Ubuntu 20.04 with a Tesla4 graphics card, using Pytorch for the entire code implementation. We sample 1000 points from the generated depth point cloud and the model point cloud. We set the epoch to 50 during the training phase. The parameters $\gamma_1=100, \gamma_2=50,\gamma_3=50$ are set in the PPF constraints term, and the parameters $\varphi_1: \varphi_2: \varphi_3: \varphi_4=1: 1: 1: 0.01$ are set in the loss function for the joint training of loss terms.

\textbf{Attention Feature Extraction.} In the process of obtaining the attention map, as shown in figure \ref{FIG:PSN}, we first apply the multilayer perceptron layer to boost the 6-dimensional data to two dimensions of 64 and 128. To retain the local features more accurately, the 64 and 128 dimensions are locally fused as local features following DenseFusion \cite{11}. Afterward, the concatenated 256-dimensional features are further encoded. Then the 1024-dimensional features are pooled on average as global features and densely fused with the previously generated local features.

\subsection{Comments on the effect of "PPF constraints"}

“PPF constraints” refer to constraining the learning of the attention network with Point Pair Features (PPF). As proposed by \cite{24}, PPF is a feature representation method used to describe the relative geometric relations between two points. \cite{24} stores the features of all points in a hash table as a global model descriptor for matching, while we only adopt the features to supervise the point-wise attention map. Specifically, we deploy the PPF based on the following principle. Two close points with similar normal vectors should achieve special PPF values, which leads to a higher weight after our aggregation formulation. Thus, the weights between the scene points after transformation and their corresponding model points should be highlighted. In that condition, it is plausible to take the PPF weights as supervision to indicate how the correspondence should be between the original model and scene points.

\subsection{Comments on the effect of the final average process of pose sets}

In the final processing of pose sets, we used the same averaging method as Bico-net \cite{43}. However, when encountering situations of inconsistent poses, such as symmetrical objects, this average calculation may lead to inaccurate pose estimation, as you mentioned. 

The characteristic of symmetric objects is that their appearance is mirror symmetric on a certain plane, so it is easy to have multiple symmetric solutions when predicting poses. When dealing with an object based on rotational axis symmetry (such as the can object in the Linemod dataset), we believe that the average operation has little impact on this type of object, as its pose remains symmetrical after averaging. However, when dealing with another object based on 180-degree symmetry (such as the eggbox object in the Linemod dataset), there may be some deviation in the final pose obtained by this average operation, resulting in poor performance of the final prediction results. 

To address this issue, we have incorporated a novel attention mechanism in our approach. The attention mechanism helps to identify and leverage the differences among predicted correspondences, effectively reducing the impact of inconsistent poses during the averaging process. By emphasizing the reliable correspondences and suppressing the less reliable ones, the attention mechanism contributes to more accurate and robust pose estimation, particularly for symmetric objects and challenging scenarios. Additionally, we are considering clustering the final pose set based on its rotation axis, and then averaging and refining the clustered pose set.

\begin{figure}[t]
    \centering
    \includegraphics[scale=.3]{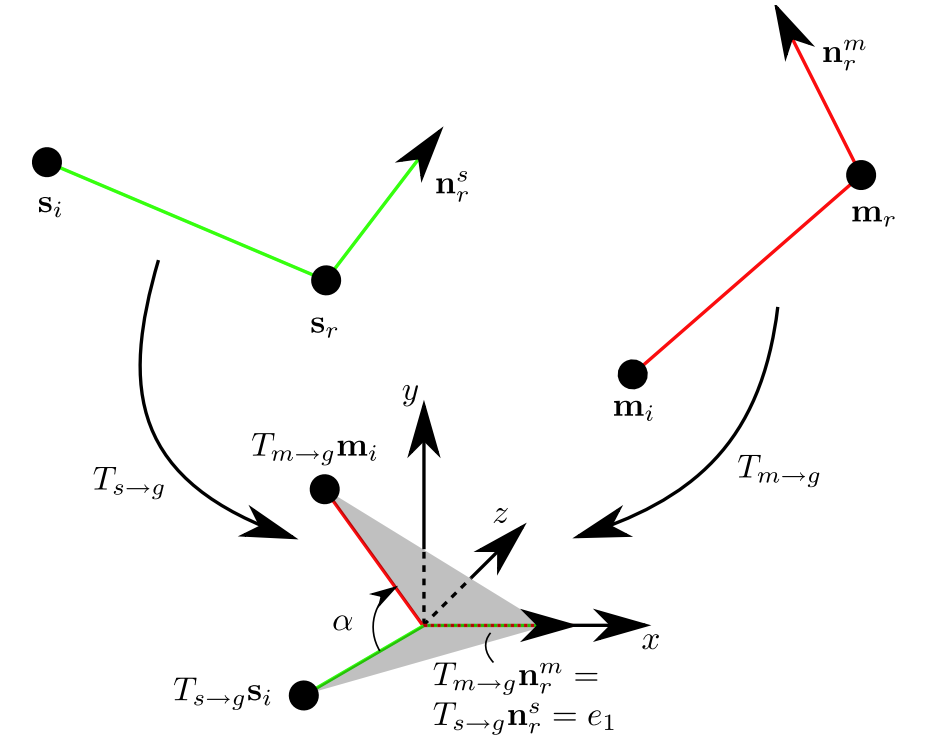}
    \caption{Transformation between model and scene coordinates in PPF \cite{24}.}
    \label{fig:ppf}
\end{figure}

\subsection{Details of the three regression pose branches}

In our method, we employ two kinds of poses, including $T_s$ and $T_m$ that are computed from the correspondence alignment branches, and $T_d$ that is predicted from the direct regression branch.

For $T_d$, we let the network predict the 3D translation vector and a normalized 4D quaternion vector.
While for $T_s$, we follow \cite{24} to define it as the transformation from a point pair $(m_r,m_i)$ in the model coordinates to its corresponding point pair $(s_r,s_i)$ in the scene coordinates
\begin{equation}
s_i=T_s m_i
\end{equation}
\begin{equation}
T_s=T_{s\rightarrow g}^{-1} R_g(\alpha) T_{m\rightarrow g}
\end{equation}

In the equation, $T_{m\rightarrow g} \in R^{4\times 4}$ transforms $m_r$ into the origin and rotates its normal vector onto the x-axis, and it is formulated by computing the 3D translation vector, 3D rotation axis, and 3D rotation angle. 
And so is the case with $T_{s\rightarrow g}$. 
$R_g(\alpha) \in SO(3)$ defines a rotation around the x-axis to align $T_{m\rightarrow g}$ and $T_{s\rightarrow g}$, which is formulated by computing the relative 3D angle.
$T_m$ is defined and computed the same as $T_s$. 
Our rotation matrices are all calculated from the rotation axis, angles, or quaternions, thus they are all originally orthonormalized. Figure \ref{fig:ppf} illustrates the process of calculating the position in PPF.

\begin{table}[tbp]
\caption{Ablation on ADD-S loss for symmetric objects}\label{sym}
\vspace{-2.5mm}
\begin{center}
\begin{tabular}{c|c|c}
\hline
\multicolumn{1}{c|}{Object} &\multicolumn{1}{p{6.1875em}|}{fine-tune with ADD loss} & \multicolumn{1}{p{7.1875em}}{with ADD-S loss} \\
\hline
eggbox & 82 &  81.3 \\
glue & 79.6 & 79  \\
\hline
\end{tabular}
\end{center}
\vspace{-6.5mm}
\end{table}

\section{More experimental results}

\subsection{Experiment against ADD-S Loss}
In our experimental results, while our network demonstrates improved accuracy on the majority of objects, we have observed that there are certain symmetric objects where the performance is not as satisfactory. To further investigate the impact of the ADD-S loss function on predicting the poses of both symmetric and asymmetric objects during the bidirectional correspondence stage, we conduct additional analyses. Specifically, We fine-tune the two symmetric objects using ADD loss and examine the results on the Occ-Linemod dataset. Table \ref{sym} presents the outcomes, indicating that the accuracy of both "glue" and "eggbox" improves after the modification of the correspondence mapping loss function. This suggests that the effectiveness of the ADD-S loss function may vary depending on the symmetry characteristics of individual objects. In the future, we will conduct more in-depth research on this aspect.

\subsection{Comparison with state-of-the-art methods on three benchmark datasets}

Table \ref{tab:linemod} shows the comparison of FFB6D\cite{12} and RCVPose\cite{rcvpose} methods on the LineMOD dataset. Table \ref{tab:ycb} shows the comparison of  FFB6D\cite{12} and RCVPose\cite{rcvpose} methods on the YCB-V dataset; and Table \ref{tab:occ} shows the comparison of FFB6D\cite{12}, RCVPose\cite{rcvpose}, CheckerPose\cite{checkerpose}, ZebraPose\cite{zebrapose}, and Depth-based 6DoF Object Pose Estimation using Swin Transformer\cite{swintf} methods on the Occ-LineMOD dataset. 

\begin{table}[t]
  \caption{The state-of-the-art works on LineMOD dataset}
  \begin{center}
    \begin{tabular}{c|c|c|c|c}
    \hline
    \multicolumn{1}{c|}{} & \multicolumn{1}{c|}{FFB6D} & \multicolumn{1}{c|}{RCVPose} & \multicolumn{1}{c|}{RCVPose+ICP} & \multicolumn{1}{c}{Ours} \\
    \hline
    ape   & 98.4  & \multicolumn{1}{c|}{-} & \multicolumn{1}{c|}{-} & 98.2 \\
    \hline
    benchvise & 100   & \multicolumn{1}{c|}{-} & \multicolumn{1}{c|}{-} & 99.7 \\
    \hline
    camera & 99.9  & \multicolumn{1}{c|}{-} & \multicolumn{1}{c|}{-} & 100 \\
    \hline
    can   & 99.8  & \multicolumn{1}{c|}{-} & \multicolumn{1}{c|}{-} & 99.8 \\
    \hline
    cat   & 99.9  & \multicolumn{1}{c|}{-} & \multicolumn{1}{c|}{-} & 100 \\
    \hline
    driller & 100   & \multicolumn{1}{c|}{-} & \multicolumn{1}{c|}{-} & 99.3 \\
    \hline
    duck  & 98.4  & \multicolumn{1}{c|}{-} & \multicolumn{1}{c|}{-} & 99 \\
    \hline
    eggbox$^\star$ & 100   & \multicolumn{1}{c|}{-} & \multicolumn{1}{c|}{-} & 99.8 \\
    \hline
    glue$^\star$  & 100   & \multicolumn{1}{c|}{-} & \multicolumn{1}{c|}{-} & 99.9 \\
    \hline
    holepuncher & 99.8  & \multicolumn{1}{c|}{-} & \multicolumn{1}{c|}{-} & 99.8 \\
    \hline
    iron  & 99.9  & \multicolumn{1}{c|}{-} & \multicolumn{1}{c|}{-} & 99.9 \\
    \hline
    lamp  & 99.9  & \multicolumn{1}{c|}{-} & \multicolumn{1}{c|}{-} & 99.8 \\
    \hline
    phone & 99.7  & \multicolumn{1}{c|}{-} & \multicolumn{1}{c|}{-} & 99.5 \\
    \hline
    MEAN  & 99.7 & 99.4 & 99.7  & 99.6 \\
    \hline
    \end{tabular}%
    \begin{tablenotes}
    \footnotesize
    \item[*] *Objects marked with stars are symmetrical objects
    \end{tablenotes}
    \end{center}
  \label{tab:linemod}%
\end{table}%

\subsection{More results on YCB-Video dataset}
In this section, we integrate the ADD-AUC metric to comprehensively assess the accuracy and robustness of our pose predictions on the YCB-Video dataset. The ADD-AUC metric, encompassing the area under the pose error curve, provides a comprehensive perspective on performance. During the evaluation, we utilize the segmentation masks provided by PVN3D \cite{34}.
Table \ref{tab:addauc} demonstrates our results that our network also possesses a significant improvement in the ADD-AUC metric compared to the current mainstream methods.

Through this supplementary section, we aim to provide a well-rounded assessment of our method's performance by incorporating the ADD-AUC metric, thereby ensuring a fair and comprehensive evaluation framework for comparing our approach with the existing state-of-the-art methods.

\begin{table}[t]
  \centering
  \caption{The state-of-the-art works on YCB-V dataset}
    \begin{tabular}{c|c|c|c|c}
    \hline
    \multicolumn{1}{r|}{} & FFB6D & RCVPose & RCVPose+ICP & \multicolumn{1}{c}{Ours} \\
    \hline
    Object & \multicolumn{1}{c|}{AUC} & \multicolumn{1}{c|}{AUC} & \multicolumn{1}{c|}{AUC} & \multicolumn{1}{c}{AUC} \\
    \hline
    002   & 96.3  & \multicolumn{1}{c|}{-} & \multicolumn{1}{c|}{-} & 95.8 \\
    \hline
    003   & 96.3  & \multicolumn{1}{c|}{-} & \multicolumn{1}{c|}{-} & 96.5 \\
    \hline
    004   & 97.6  & \multicolumn{1}{c|}{-} & \multicolumn{1}{c|}{-} & 98.1 \\
    \hline
    005   & 95.6  & \multicolumn{1}{c|}{-} & \multicolumn{1}{c|}{-} & 95.9 \\
    \hline
    006   & 97.8  & \multicolumn{1}{c|}{-} & \multicolumn{1}{c|}{-} & 98.3 \\
    \hline
    007   & 96.8  & \multicolumn{1}{c|}{-} & \multicolumn{1}{c|}{-} & 97.1 \\
    \hline
    008   & 97.1  & \multicolumn{1}{c|}{-} & \multicolumn{1}{c|}{-} & 98.1 \\
    \hline
    009   & 98.1  & \multicolumn{1}{c|}{-} & \multicolumn{1}{c|}{-} & 98.7 \\
    \hline
    010   & 94.7  & \multicolumn{1}{c|}{-} & \multicolumn{1}{c|}{-} & 93.4 \\
    \hline
    011   & 97.2  & \multicolumn{1}{c|}{-} & \multicolumn{1}{c|}{-} & 97.6 \\
    \hline
    019   & 97.6  & \multicolumn{1}{c|}{-} & \multicolumn{1}{c|}{-} & 97.6 \\
    \hline
    021   & 96.8  & \multicolumn{1}{c|}{-} & \multicolumn{1}{c|}{-} & 96.8 \\
    \hline
    024$^\star$ & 96.3  & \multicolumn{1}{c|}{-} & \multicolumn{1}{c|}{-} & 96 \\
    \hline
    025   & 97.3  & \multicolumn{1}{c|}{-} & \multicolumn{1}{c|}{-} & 97.4 \\
    \hline
    035   & 97.2  & \multicolumn{1}{c|}{-} & \multicolumn{1}{c|}{-} & 97.3 \\
    \hline
    036$^\star$ & 92.6  & \multicolumn{1}{c|}{-} & \multicolumn{1}{c|}{-} & 94.1 \\
    \hline
    037   & 97.7  & \multicolumn{1}{c|}{-} & \multicolumn{1}{c|}{-} & 93.5 \\
    \hline
    040   & 96.6  & \multicolumn{1}{c|}{-} & \multicolumn{1}{c|}{-} & 98 \\
    \hline
    051$^\star$ & 96.8  & \multicolumn{1}{c|}{-} & \multicolumn{1}{c|}{-} & 92 \\
    \hline
    052$^\star$ & 96    & \multicolumn{1}{c|}{-} & \multicolumn{1}{c|}{-} & 86.9 \\
    \hline
    061$^\star$ & 97.3  & \multicolumn{1}{c|}{-} & \multicolumn{1}{c|}{-} & 96.9 \\
    \hline
    MEAN & 96.6  & 96.6  & 97.2  & 95.8 \\
    \hline
    \end{tabular}%
    \begin{tablenotes}
    \footnotesize
    \item[*] *Objects marked with stars are symmetrical objects
\end{tablenotes}
  \label{tab:ycb}%
\end{table}%

\begin{table*}[h]
  \centering
  \caption{The state-of-the-art works on Occ-LineMOD dataset}
  \begin{center}
\begin{threeparttable}
    \begin{tabular}{c|c|c|c|c|c|c}
    \hline
    \multicolumn{1}{c|}{} & \multicolumn{1}{c|}{FFB6D} & \multicolumn{1}{c|}{RCVPose+ICP} & \multicolumn{1}{c|}{ CheckerPose} & \multicolumn{1}{c|}{Zebrapose} & \multicolumn{1}{c|}{swim transformer}  & \multicolumn{1}{c}{Ours} \\
    \hline
    ape   & 47.2  & \multicolumn{1}{c|}{-} & 58.3  & 57.9  & 59.8    & 58.3 \\
    \hline
    can   & 85.2  & \multicolumn{1}{c|}{-} & 95.7  & 95    & 88.8    & 88.5 \\
    \hline
    cat   & 45.7  & \multicolumn{1}{c|}{-} & 62.3  & 60.6  & 46.7    & 51.6 \\
    \hline
    driller & 81.4  & \multicolumn{1}{c|}{-} & 93.7  & 94.8  & 95.1   & 77.8 \\
    \hline
    duck  & 53.9  & \multicolumn{1}{c|}{-} & 69.9  & 64.5  & 59.4   & 64.8 \\
    \hline
    eggbox$^\star$ & 70.2  & \multicolumn{1}{c|}{-} & 70    & 70.9  & 90.3    & 81.3 \\
    \hline
    glue$^\star$  & 60.1  & \multicolumn{1}{c|}{-} & 86.4  & 88.7  & 88    & 79 \\
    \hline
    holepuncher & 85.9  & \multicolumn{1}{c|}{-} & 83.8  & 83    & 88.8   & 93.6 \\
    \hline
    MEAN  & 66.2  & 71.1  & 77.5  & 76.9  & 77.1   & 74.4 \\
    \hline
    \end{tabular}
    \begin{tablenotes}
    \footnotesize
    \item[*] Objects marked with stars are symmetrical objects
    \end{tablenotes}
    \end{threeparttable}
    \end{center}
  \label{tab:occ}%
\end{table*}

\begin{table*}[t]
  \centering
  
  \caption{Evaluation results in terms of the ADD(AUC) evaluation metric on YCB-Video dataset}
  \begin{threeparttable}
  
    \begin{tabular}{c|c|c|c|c|c|c}
    \hline
    \multicolumn{1}{c|}{} & \multicolumn{1}{c|}{DenseFusion} & \multicolumn{1}{c|}{PoseRBPF} & \multicolumn{1}{c|}{PVN3D} & \multicolumn{1}{c|}{PVN3D+ICP} & \multicolumn{1}{c|}{BiCo-net} & \multicolumn{1}{c}{Ours} \\
    \hline
    Object & \multicolumn{1}{c|}{ADD-AUC} & \multicolumn{1}{c|}{ADD-AUC} & \multicolumn{1}{c|}{ADD-AUC} & \multicolumn{1}{c|}{ADD-AUC} & \multicolumn{1}{c|}{ADD-AUC} & \multicolumn{1}{c}{ADD-AUC} \\
    \hline
    002   & 70.7  & \textbf{91.9} & 80.5  & 79.3  & 79.2  & 79.8 \\
    \hline
    003   & 86.9  & 91.8  & 94.8  & 91.5  & 94.9  & \textbf{95.8} \\
    \hline
    004   & 90.8  & 94    & 96.3  & 96.9  & 96.7  & \textbf{97.4} \\
    \hline
    005   & 84.7  & 91    & 88.5  & 89    & 89.7  & \textbf{92} \\
    \hline
    006   & 90.9  & 93.2  & 96.2  & 97.9  & 97.3  & \textbf{97.9} \\
    \hline
    007   & 79.6  & 80    & \textbf{89.3} & 90.7  & 78.9  & 84.5 \\
    \hline
    008   & 89.3  & 80.6  & 95.7  & 97.1  & 96.2  & \textbf{97.3} \\
    \hline
    009   & 95.8  & 96.4  & 96.1  & 98.3  & 97.6  & \textbf{98.3} \\
    \hline
    010   & 79.6  & 77.8  & \textbf{88.6} & 87.9  & 86.3  & 87.2 \\
    \hline
    011   & 76.7  & 87.5  & 93.7  & 96    & 93.1  & \textbf{96.3} \\
    \hline
    019   & 87.1  & 89.8  & 96.5  & 96.9  & 95.9  & \textbf{97.5} \\
    \hline
    021   & 87.5  & 88.6  & 93.2  & \textbf{95.9} & 94.4  & 94.9 \\
    \hline
    024$^\star$ & 86    & 46.8  & 90.2  & 92.8  & \textbf{96.5} & 92 \\
    \hline
    025   & 83.8  & 91.4  & 95.4  & \textbf{96} & 88.9  & 91.8 \\
    \hline
    035   & 83.7  & 95.1  & 95.1  & 95.7  & 94.7  & \textbf{96.7} \\
    \hline
    036$^\star$ & 89.5  & 33.4  & 90.4  & 91.1  & \textbf{95.2} & 94.3 \\
    \hline
    037   & 77.4  & 89    & \textbf{92.7} & 87.2  & 82.6  & 78.4 \\
    \hline
    040   & 89.1  & 91.6  & 91.8  & 91.6  & 91.5  & \textbf{92.4} \\
    \hline
    051$^\star$ & 71.5  & 90.9  & 93.6  & 95.6  & \textbf{95.9} & 93.3 \\
    \hline
    052$^\star$ & 70.2  & 77    & 88.4  & 90.5  & 95.1  & \textbf{95.9} \\
    \hline
    061$^\star$ & 92.2  & 95.3  & 96.8  & \textbf{98.2} & 96.8  & 96.4 \\
    \hline
    \textbf{ALL} & 82.9  & 86.8  & 91.8  & 92.3  & 91.2  & \textbf{92.3} \\
    \hline
    \end{tabular}
    \begin{tablenotes}
    \footnotesize
    \item[*] Objects marked with stars are symmetrical objects
\end{tablenotes}
    \end{threeparttable}
  \label{tab:addauc}%
\end{table*}%

From Table \ref{tab:linemod}, Our method achieves a competitive accuracy of 99.6\%, closely trailing the leading method by only 0.1\%. The key advantage of our approach is achieving comparable performance to methods utilizing synthetic and fuse data (FFB6D), while relying solely on real-world data for training. This ensures generalizability and robustness in handling diverse and challenging scenarios without the need for costly synthetic data generation.
Additionally, our method's efficiency and simplicity, without iterative optimization, make it practical for real-time applications in robotics and augmented reality systems. 
By leveraging only real data, our approach offers a compelling choice for practical 6D pose estimation applications.

From the results in Table \ref{tab:ycb} and \ref{tab:occ}, it is evident that our method may not be the top-performing one on specific datasets compared to other state-of-the-art approaches. However, these observations suggest potential areas for optimization within our network, such as refining calculation methods for symmetric objects in pose regression and enhancing the performance of direct regression branches.

Taking into account the overall effectiveness of our network across multiple datasets, our experimental results demonstrate its exceptional performance beyond the limitations of the baseline. Consequently, we firmly believe that the primary advantage of our network lies in its ability to significantly enhance 6D pose estimation across diverse datasets, positioning it as a current state-of-the-art solution in this domain. This capability makes our method a compelling choice for practical applications where generalization and robustness are crucial factors.

\section{Contributions and Expansion of Our Approach}
\subsection{Analysis of applicability to other frameworks}
First, the proposed attention map is validated to be helpful for both correspondence prediction and direct regression pipelines.Thanks to the combination of both pipelines in BiCo-Net \cite{43}, we are enabled to verify the effectiveness of the attention mechanism respectively, as shown in Table \ref{tbl4}. 

As for other methods besides BiCo-Net, we argue that similar processes, i.e. predicting correspondence or poses from encoded features, could all benefit from the extra attention information more or less.
PVNet and PVN3D differ from BiCo-Net in predicting sparse keypoints locations rather than dense matching points.
However, fundamentally, they are still learning the corresponding points between the scene and the model from the encoded features.
Therefore, the attention maps between scene and model points could also help the learning process.

Second, the proposed attention mechanism is highly versatile and adaptive to other architectures. 
The input of our module, the scene and the model points, is commonly applicable in RGBD-based instance-level pose estimation tasks. 
And the output of our module, the attention maps, can be simply concatenated with the original features.
For example, if we were to use this module on PVN3D, we could concatenate the learned attention maps to the N*1792 feature vector in PVN3D.
In future work, we are more than happy to conduct such in-depth research in this perspective.

\subsection{Our contribution and improvements}
BiCo-Net proposes a novel bidirectional correspondence prediction network to further exploit the CAD
model information, and uniquely combines global regression and local matching for robust 6D pose estimation.
It is capable of handling challenging scenarios, such as occlusion and clutters, and achieving great performance.
Based on the main architecture of BiCo-Net, our contributions lie in proposing a global point-wise attention mechanism to leverage the similarities between observations and the model prior. Experiments and ablation studies show the effectiveness of our proposed attention mechanism.

\end{appendices}

\end{document}